\documentclass{article}
\ifdefined\pdfminorversion
\fi
\PassOptionsToPackage{numbers,compress}{natbib}
\usepackage[eandd,final]{neurips_2026}

\usepackage[utf8]{inputenc}
\usepackage[T1]{fontenc}
\usepackage{hyperref}
\usepackage{url}
\usepackage{booktabs}
\usepackage{amsfonts}
\usepackage{nicefrac}
\usepackage{microtype}
\usepackage{xcolor}

\usepackage{graphicx}
\usepackage{subcaption}
\usepackage{enumitem}
\usepackage{multirow}
\usepackage{wrapfig}
\usepackage{bm}
\usepackage{makecell}
\usepackage{colortbl}
\usepackage{array}
\usepackage{nicematrix}
\usepackage{adjustbox}
\usepackage{xspace}
\usepackage{needspace}

\usepackage{amsmath,amssymb,amsthm}
\usepackage{mathtools}
\usepackage[capitalize,noabbrev]{cleveref}

\theoremstyle{plain}

\theoremstyle{definition}

\theoremstyle{remark}

\providecommand{\tt}{\mathbf{t}}

\newcommand{\eg}{\emph{e.g.}\xspace}

\newcolumntype{M}[1]{>{\centering\raggedright\arraybackslash}m{#1}}

\title{MEV: A Multi-Event Video Dataset for Long-Take Generation}

\author{%
  \textbf{Peiyuan Zhu}$^{1}$ \quad
  \textbf{Shaoan Xie}$^{2,1}$ \quad
  \textbf{Yifan Shen}$^{1}$ \quad
  \textbf{Wen Tian}$^{2}$ \\
  \textbf{Zijian Li}$^{1,2}$ \quad
  \textbf{Guangyi Chen}$^{1,2}$\thanks{Corresponding author.} \quad
  \textbf{Kun Zhang}$^{2,1}$ \\[0.3em]
  $^{1}$Mohamed bin Zayed University of Artificial Intelligence \\
  $^{2}$Carnegie Mellon University \\[0.4em]
  {\normalfont\small Project page: %
    \hypersetup{hidelinks}%
    \href{https://mev-dataset.github.io/}{\textcolor{blue!60!black}{\texttt{https://mev-dataset.github.io/}}}}
}

\begin{document}

\maketitle

\raggedbottom

\vspace{-1em}
\begin{figure}[!ht]
  \centering
   \includegraphics[width=\linewidth]{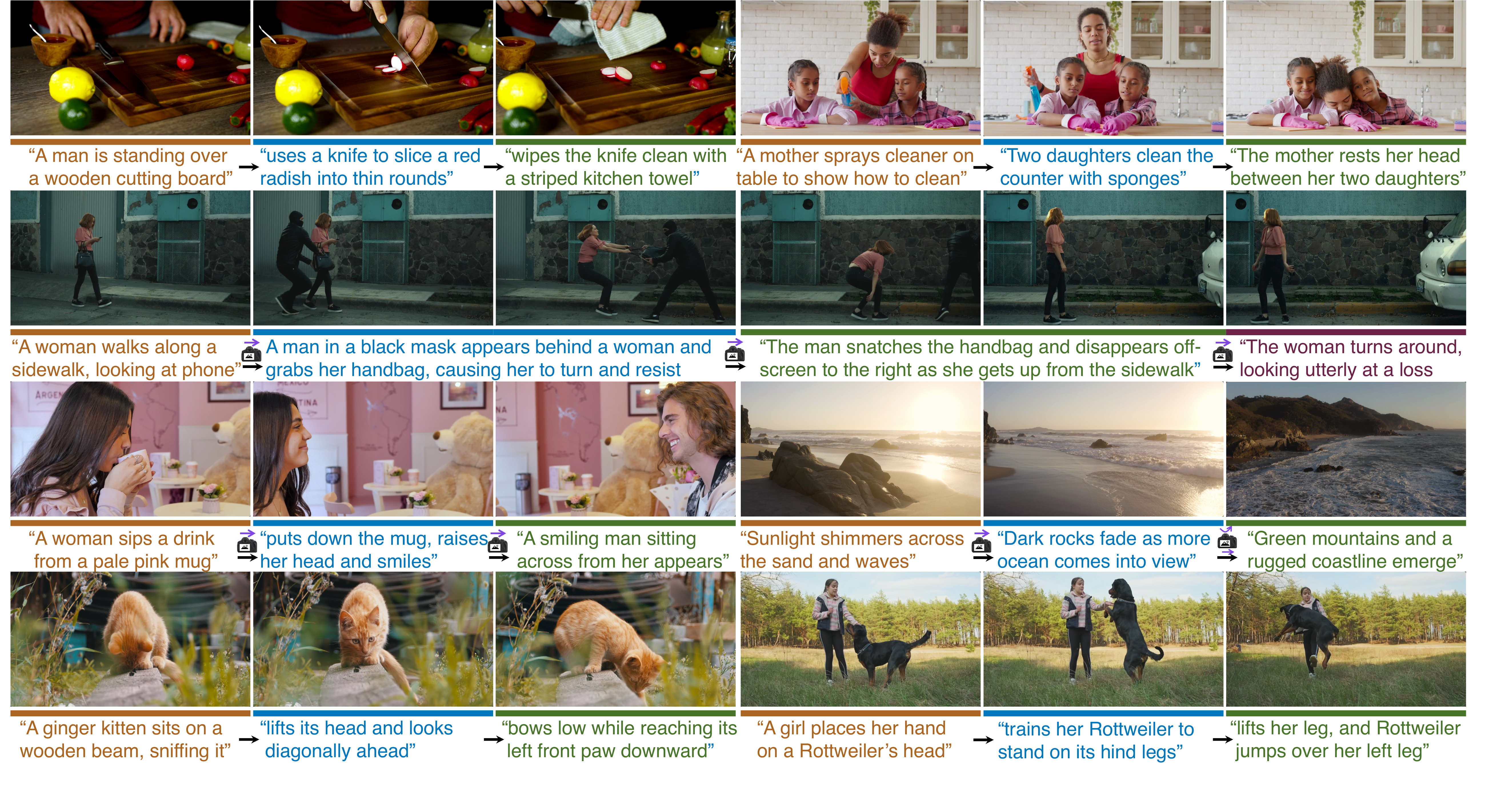}
   \vspace{-1.5em}
   \caption{Examples of multi-event narratives in the MEV dataset. MEV contains high-quality long-take videos segmented into events with precise start/end timestamps and fine-grained procedural descriptions. Each video combines global–event hierarchical captions with temporally aligned camera annotations and semantic attributes, providing structured supervision for multi-event generation.}
   \label{fig:test}
\end{figure}

\begin{abstract}
Current datasets for text-to-video generation are largely limited to short clips depicting a single event or coarse-grained captions. As a result, they provide limited supervision for modeling long-take temporal dependencies, often leading to temporally inconsistent generations when describing multiple events. In this paper, we introduce MEV, a multi-event video dataset designed for long-take video generation. MEV provides high-quality real-world videos where subjects can perform a sequence of distinct, non-overlapping events, featuring precise temporal boundaries and fine-grained procedural annotations. This event-centric discretization reduces confounding factors and enables models to support temporally localized conditioning and build long-take videos from atomic event units. Furthermore, we introduce a comprehensive annotation pipeline that segments videos into temporally grounded events, and generates structured captions at both the global level and the event level. In addition, under an event-centric evaluation protocol, we conduct multi-event generation experiments by fine-tuning on MEV with temporally grounded event conditioning, leading to improved long-take event transitions and more coherent content. We believe this work provides a publicly available resource for multi-event fine-tuning and event-centric evaluation, together with practical insights into learning from temporally structured video--text data.
\end{abstract}
\section{Introduction}
\begin{quote}
\raggedright 
{\em ``The world we know is a continuous stream of occurrence which we can discriminate into finite events~\ldots''}\\
\hfill--- {\scriptsize Alfred North Whitehead, \emph{The Concept of Nature}~\citeyearpar{whitehead1920concept}}
\end{quote}

An event is a segment of time at a given location that is conceived by an observer to have a beginning and an end~\citep{zacks2007event}. The human brain inherently possesses an “Event Segmentation” mechanism, allowing us to automatically delineate the onset and offset of actions within the seamless flow of reality~\citep{james1890principles,zacks2007event}. While this recognition is intuitive for humans, it remains a formidable challenge for video models to align continuous pixel streams with discrete semantic boundaries. Closing this gap is pivotal for mastering video intelligence. This pursuit requires a paradigm shift from mere architectural exploration toward the cultivation of training signals that capture the discrete event structures inherent in visual data.

In realistic scenarios, even simple activities unfold as a sequence of distinct but related events: a person may open a fridge, take out a bottle, walk to a table, pour a drink, and then sit down and read. Generative models that aim to serve as general-purpose video creators must therefore be able to synthesize such multi-event narratives, respecting both local motion realism and the global structure imposed by the sequence of actions. Current models can generate visually plausible long videos through a variety of paradigms, ranging from multi-shot composition~\citep{cai2025mixture, guo2025long, wang2025multishotmaster} to long-take, single-shot sequences that exhibit scarce action~\citep{henschel2025streamingt2v, cai2025ditctrl}. 
Between these two extremes lies a less-explored but critical challenge: the generation of long-take multi-event narratives, which requires maintaining an unbroken camera perspective while navigating through a series of semantically evolving activities.
While some methods~\citep{yang2025longlive, lu2025reward, po2025bagger} have been developed to facilitate long-take multi-event generation, they frequently encounter unstable drifting and illogical event transitions as the sequence progresses.

Datasets play a crucial role in enabling video generation models to understand and generate such multi-event structures. However, existing video–text datasets exhibit persistent limitations. Many datasets are dominated by trimmed clips with limited action richness~\citep{chen2024panda,tan2024vidgen,nan2024openvid,wang2025koala,bolya2025perception}, while explicit temporal boundaries—crucial supervision signals for modeling long-take multi-event videos under divide-and-conquer or temporal autoregressive paradigms—are often missing. Even with temporal boundaries, datasets may suffer from temporal discontinuities, cluttered scenes with highly intertwined activities and scene cuts, and captions that remain coarse-level—lacking critical stylistic and temporal nuances such as lighting, camera dynamics, and fine-grained action details—rather than specifying concrete, stepwise actions~\citep{Heilbron2015ActivityNetAL,oncescu2021queryd,soldan2022mad,chen2024mecd,huang2020movienet,wang2025videoufo}. This multifaceted mismatch creates a systemic bottleneck that precludes the development of models for long-take, compositionally controlled generation.

To fill these gaps, we propose MEV, a high-quality, publicly released dataset specifically designed to advance the technical frontiers of long-take multi-event video generation. MEV comprises more than 8k high-definition (720p–4k) long videos, with each sequence encompassing between 2 and 21 distinct events, totaling over 25k event-level video segments. From a global perspective, the dataset provides multi-granularity captions for the entire video, supplemented by 8 structured captions such as lighting and subject profile; at the local level, each event is paired with a caption, camera annotations, semantic attributes, and VBench quality scores, which provide interpretable supervision for modeling dynamic changes and enable optional slicing for subset selection and analysis. All events are precisely grounded with clear start and end timestamps relative to the original long-take video, making MEV, to the best of our knowledge, the first public dataset to jointly provide such structured, global-local hierarchical captions for long-take video generation. To obtain this dataset, we conduct a rigorous four-stage data curation process: (1) video collection of high-resolution content from public internet sources; (2) filter out non-conforming clips via statistical filtering and model-based purification; (3) segment videos into non-overlapping events with rule-guided atomic grounding for precise boundaries; (4) annotate global–event structured captions with multiple granularities, plus event-level camera descriptions, semantic attributes and quality scores. We conduct multi-perspective statistical analyses of MEV to comprehensively characterize the dataset, and perform evaluations on multi-granular captions, revealing potential limitations of existing video models in capturing event structure. Furthermore, we validate MEV on long-take multi-event generation by finetuning Wan 2.1~\citep{wan2025wan} on MEV with temporally grounded event conditioning, showing improved event transitions and more coherent long-take narratives.
In summary, our contributions are threefold:
\begin{itemize}[leftmargin=12pt,topsep=3pt,itemsep=2pt,parsep=0pt]
  \item To support the rapidly advancing multi-event video generation applications and bridge the gap between academic datasets and large corporate data, we build MEV, a high-quality video dataset with precisely grounded event boundaries and fine-grained structured captions, focusing on fine-tuning fundamental video generation models with grounded event conditioning.
  \item We propose a comprehensive annotation scheme with global- and event-level structured captions, camera-aware descriptions, and fine-grained semantic attributes, providing temporal grounding for complex video–text alignment.
  \item We introduce MEV-Eval, a supporting event-centric evaluation protocol for measuring event execution, order, temporal grounding, and boundary transition coherence. We conduct long-take multi-event video generation experiments by finetuning Wan2.1 on MEV with temporally grounded event conditioning, showing that MEV supports coherent multi-event storytelling and consistently outperforms the counterpart dataset across evaluations.
\end{itemize}

\vspace{-0.1em}
\section{MEV Dataset}
\label{sec:dataset}
\begin{figure*}[t]
  \centering
   \includegraphics[width=\linewidth]{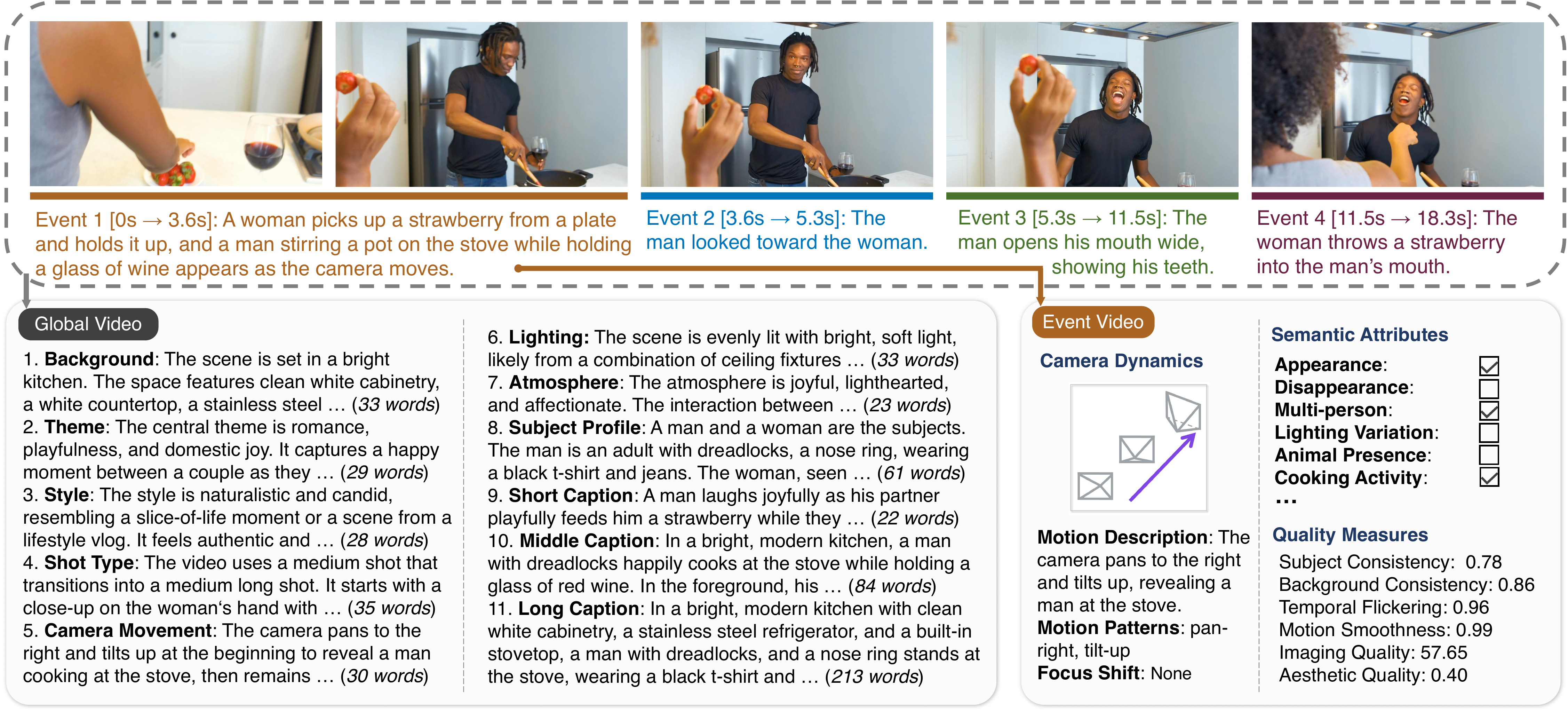}
   \caption{Comprehensive annotation scheme of the MEV dataset. MEV provides video-level captions at multiple granularities (short/middle/long), complemented by a structured set of global fields (e.g., background, theme, style, shot type, camera movement, lighting, atmosphere, and subject profile).
At the event level, each long-take video is temporally segmented into atomic events with precise timestamps under a consistent camera perspective, and each event is annotated with camera-based descriptions, semantic attributes for subset selection or conditional supervision, and VBench-based quality scores assessing generation-relevant video quality across six dimensions. }
   \label{fig:scheme}
\end{figure*}
MEV is a long-take multi-event text-video dataset featuring precise temporal grounding and comprehensive global-event level captions. We illustrate the annotation scheme of MEV dataset in Fig.~\ref{fig:scheme}. The dataset is constructed through a four-step process: data collection \& curation (Sec.~\ref{mev_step_12}), event segmentation (Sec. \ref{mev_step_3}), and comprehensive annotation (Sec. \ref{mev_step_4}). In Sec.~\ref{mev_anal}, we provide a statistical analysis of the MEV dataset and a comparison with existing representative text-video datasets.

\begin{table*}[tp]
    \centering
    \caption{Comparison of MEV and popular text-to-video datasets. MEV provides high-resolution videos with a comprehensive annotation suite. Checkmarks ($\checkmark$) indicate the presence of a specific feature, while hyphens (--) denote its absence. For MEV, Global averages full original-video durations; Event averages annotated interval durations, whose sum is reported in Table~\ref{tab:mev_split}.}
    \label{tab:dataset}
    \renewcommand{\arraystretch}{1.3}
    \setlength\tabcolsep{2.0pt}
    \resizebox{1.0\linewidth}{!}{
        \begin{tabular}{lcccccccccc}
        \toprule[0.1em]
        Dataset & Resolution & \begin{tabular}[c]{@{}c@{}}Structured\\Caption\end{tabular} & \begin{tabular}[c]{@{}c@{}}Temporal\\Groundings\end{tabular} & \begin{tabular}[c]{@{}c@{}}Multi-\\granularity\end{tabular} & \begin{tabular}[c]{@{}c@{}}Camera\\Perspective\end{tabular} & \begin{tabular}[c]{@{}c@{}}Semantic\\Attributes\end{tabular} & \begin{tabular}[c]{@{}c@{}}Quality\\Scores\end{tabular} & \begin{tabular}[c]{@{}c@{}}Avg. Caption\\Length\end{tabular} & \begin{tabular}[c]{@{}c@{}}Avg. Video\\Length\end{tabular} & \begin{tabular}[c]{@{}c@{}}Video\\Clips\end{tabular} \\
        \hline
        HowTo100M~\citep{miech2019howto100m} & 240p & -- & -- & -- & -- & -- & -- & 4.0 words & 3.6s & 136M \\
        WebVid-10M~\citep{bain2021frozen} & 360p & -- & -- & -- & -- & -- & -- & 12.0 words & 17.5s & 10M \\
        HD-VILA-100M~\citep{xue2022advancing} & 720p & -- & -- & -- & -- & -- & -- & 32.5 words & 13.4s & 103M \\
        InternVid~\citep{wang2023internvid} & 720p & -- & -- & -- & -- & -- & -- & 17.6 words & 11.7s & 234M \\
        Panda-70M~\citep{chen2024panda} & 720p & -- & -- & -- & -- & -- & -- & 13.2 words & 8.5s & 70.8M \\
        MiraData~\citep{ju2024miradata} & 720p & $\checkmark$ & -- & -- & -- & -- & -- & 318.0 words & 72.1s & 330K \\
        VIDGEN-1M~\citep{tan2024vidgen} & 720p & -- & -- & -- & -- & -- & -- & 89.3 words & 10.6s & 1M \\
        LVD-2M~\citep{xiong2024lvd} & 720p & -- & -- & $\checkmark$ & -- & -- & -- & 88.8 words & 20.2s & 2.1M \\
        Koala-36M~\citep{wang2025koala} & 720p & $\checkmark$ & -- & -- & -- & -- & $\checkmark$ & 202.3 words & 13.6s & 36M \\
        OpenSoraPlan~\citep{lin2024open} & 1080p & -- & -- & -- & -- & -- & -- & 100.2 words & 20.1s & 512K \\
        OpenVid-1M~\citep{nan2024openvid} & 720p & -- & -- & -- & -- & -- & $\checkmark$ & 126.5 words & 7.2s & 1M \\
        OpenVidHD-0.4M~\citep{nan2024openvid} & 1080p & -- & -- & -- & -- & -- & $\checkmark$ & 104.5 words & 9.6s & 433K \\
        VideoUFO~\citep{wang2025videoufo} & 720p & $\checkmark$ & $\checkmark$ & $\checkmark$ & -- & -- & $\checkmark$ & 155.5 words & 12.6s & 1M \\
        UltraVideo-Long~\citep{xue2025ultravideo} & 4K / 8K & $\checkmark$ & -- & -- & $\checkmark$ & -- & -- & 850.3 words & 30.9s & 17K \\
        ViMix-14M~\citep{yang2025vimix} & 240p--1080p & -- & $\checkmark$ & $\checkmark$ & -- & -- & $\checkmark$ & 174.8 words & 6.0s & 13.7M \\
        \hline
        MEV (Ours) & 720p -- 4K & $\checkmark$ & $\checkmark$ & $\checkmark$ & $\checkmark$ & $\checkmark$ & $\checkmark$ & \begin{tabular}[c]{@{}c@{}} Global: 652.4 \\ Event: 77.5 \end{tabular} & \begin{tabular}[c]{@{}c@{}} Global: 16.7s \\ Event: 5.3s \end{tabular} & \begin{tabular}[c]{@{}c@{}} 8K \\ 25K \end{tabular} \\
        \hline
        \bottomrule[0.1em]
        \end{tabular}
    }
\end{table*}
\subsection{Data Collection and Curation}
\label{mev_step_12}
Video sources largely determine the dataset's content distribution, motion patterns, and the upper bound of visual quality. We therefore enforce strict source-level selection---prioritizing high-quality providers and carefully chosen content pools---to limit the number of low-quality candidates entering downstream filtering. Since MEV targets long-take multi-event narratives in real-world settings, we collect candidate videos with the following priorities: (i) real-world footage with rich and evolving activities, (ii) sufficient resolution and aesthetic quality, (iii) long-take characteristics with meaningful camera motion/viewpoint changes, and (iv) diverse subjects (\eg, different genders, ages, and appearances), as well as animals and nature scenes. We gather videos primarily from MixKit~\citep{mixkit_website} and complement them with YouTube videos, resulting in 227K raw video clips as the starting pool for subsequent curation. MixKit offers curated stock videos with stable imaging quality and strong aesthetics (720p--4K), while YouTube complements long-tail human-centric activities and motion patterns; to avoid noisy crawling, we only keep videos that showcase human motion (sports/fitness, cooking, daily activities, DIY/tool-use) and are at least 720p, and then curate videos in a coarse-to-fine manner from the following two aspects:

\noindent \textbf{Statistical data filtering.}
We apply a set of fast, frame-sampled checks with hard thresholds to remove clips with non-natural visual signals, using conservative criteria to avoid unnecessarily discarding valid long-take videos. Specifically, we filter videos based on: (i) overlay coverage, where we estimate the area ratio of detected on-screen text/graphics and remove clips with $>3\%$ coverage for more than $12\%$ of sampled frames; (ii) letterbox/border ratio, where persistent dark boundary bands indicate edited footage and we discard clips if this artifact appears in more than $8\%$ of frames; (iii) illumination extremes, where under-/over-exposed frames (based on luminance statistics) exceed $10\%$ of frames; and (iv) color collapse, where low-chroma frames indicating near-grayscale appearance exceed $18\%$ of frames. At this stage, we obtain 87K clips.

\noindent \textbf{Model-based purification.}
We progressively apply models from lightweight, task-specific detectors to general-purpose MLLMs for higher-level purification. We first use TransNet v2~\citep{soucek2024transnet} to detect shot transitions and remove clips with hard cuts or frequent transitions that violate the long-take setting. We then compute temporal motion scores from RAFT optical flow~\citep{teed2020raft} to summarize motion magnitude and temporal stability, and remove clips that are either near-static or dominated by abrupt, jitter-like motion. Finally, we use Qwen3-VL-32B~\citep{Qwen3-VL} to flag clips that are misaligned with MEV's target setting, and only discard videos when the model returns high-confidence violations, including: (i) overly many individuals (we allow up to three primary subjects), (ii) excessively complex scene changes caused by frequent camera movement and rapid viewpoint switching (\eg, moving-room property tours), (iii) overly macroscopic scenes with dense, small-scale motions (\eg, ports with many vessels), and (iv) trivially repetitive motions with little semantic progression (\eg, continuous push-ups). This purification step yields 41K candidate clips for event triage.

\subsection{Event Segmentation}
\label{mev_step_3}
MEV targets multi-event sequences, thus we must decide whether each curated video contains multiple events and, if so, provide precise temporal boundaries.

\noindent \textbf{Event definition: from semantic stability to process-driven atomicity.}
In video understanding and related tasks~\citep{chen2024mecd}, an event is often defined as a temporally contiguous segment in which the target/intent, interaction object, and scene state are relatively stable, and boundaries correspond to a state change rather than merely a change in motion pattern. For example, a man making a thumbs-up gesture and then a heart gesture is frequently perceived as a single event (gesturing), and a sequence like lifting a cup, sipping, and putting it back is commonly summarized as drinking.
While such coarse segmentation preserves macro semantics, it can obscure the fine-grained procedural structures that are crucial for text-to-video generation, particularly for rich-action following and controllability. Therefore, MEV adopts a more process-driven view: we treat an event as an action unit that (i) revolves around the same micro-intent, (ii) operates on the same interaction object (or the same object set), and (iii) can be described by a short natural-language procedural phrase. This design intentionally encourages action-driven rather than purely state-driven segmentation, providing more informative intermediate supervision for multi-event storytelling.

\noindent \textbf{Event Transition Signals.}
We consider event boundaries to be triggered by either \emph{state transitions} or \emph{state continuations} (i.e., stage completion that naturally closes a unit), both of which serve as intermediate checkpoints for action-rich multi-event generation and help reduce drifting and illogical transitions in long rollouts. In practice, boundary cues include:
\begin{itemize}[leftmargin=16pt,labelwidth=7.5pt,labelsep=4pt,align=left]
\item \textbf{Intent change:} \eg, shifting from "searching for an item" to "making a phone call."
\item \textbf{Interaction object change:} \eg, switching focus from "talking to person A" to "person B," or from "holding a cup" to "picking up keys."
\item \textbf{Functional scene transition:} Transitions between areas (kitchen to living room) or stages (preparation to execution).
\item \textbf{Semantic outcome change:} The completion or failure of a task, or the start of a subsequent logical step.
\item \textbf{Salient pause or turning point:} Significant pauses, the end of a salient motion episode, or camera pans to new regions.
\end{itemize}

\noindent \textbf{Rule-based Atomic Grounding.}
Even advanced MLLMs~\citep{Comanicietal2025,Qwen3-VL} can struggle with precise temporal alignment, collapsing multiple events into one or producing boundaries that mismatch human perception. We thus develop a three-step rule-guided atomic grounding scheme for manual labeling.

\noindent\textit{\textbf{Step 1}: Multi-event Discernment.}
We first determine if a video qualifies as a multi-event sequence. It must exhibit \textbf{separability} (at least two independently nameable sub-actions) and \textbf{reproducibility} (boundaries must be stable across different annotators, i.e., most annotators place them at similar moments). 
A video is labeled as \emph{single-event} and is filtered out if (1) a single dominant activity persists throughout (micro-intent unchanged; object changes do not constitute a new action unit), and (2) any further splitting only yields local fragments of the same action with little naming benefit.
We label a video as \emph{difficult} and filter it out when candidate events cannot be consistently named or their boundaries cannot be reproducibly localized. Sports and performance videos can be particularly challenging when continuous motion obscures distinct action stages. We nevertheless retain such videos when they satisfy the same separability and reproducibility criteria, allowing continuous motion within each event.

\noindent\textit{\textbf{Step 2}: Boundary Localization.}
For videos labeled as multi-event, boundaries must align with the start/completion of action units.
When two candidate sub-actions are fully continuous and lack a reproducible cut point (no pause and no clear procedural result), we allow merging them into a single event (\eg, ``pick up a cup and drink'' as one unit).

\noindent\textit{\textbf{Step 3}: Empirical Refinement Rules.}
To ensure consistency and engineering utility, we apply four "merging and filtering" rules:
\begin{enumerate}[leftmargin=16pt,labelwidth=7.5pt,labelsep=4pt,align=left,itemsep=0pt,parsep=0pt]
\item \textbf{Continuity Rule:} Closed-loop actions (\eg, "reaching for a pen" and "writing") are merged into a single segment if no stable pause exists.
\item \textbf{Coordination Rule:} If multiple body parts cooperate concurrently to accomplish the same action unit, keep them in the same segment unless one part is clearly emphasized as the main action (\eg, a close-up of a hand).
\item \textbf{Camera Motion Rule:} Camera tracking alone does not trigger a new event. A change in camera behavior can define a new event when it forms an independently nameable visual stage with reproducible temporal boundaries.
\item \textbf{Incidental Motion Rule:} Unconscious auxiliary motions (\eg, hand gestures while talking) are merged into the primary event unless explicitly emphasized by the composition.
\end{enumerate}

\noindent \textbf{Annotation protocol and scalability.}
We employ three annotators to label 52.9\% of videos from the curated pool, providing event boundaries (or marking them as single-event/difficult) and performing cross-checks to ensure reproducibility.
To tolerate ambiguity in transition regions, for any boundary timestamps $t_1$ and $t_2$ from different annotators we treat them as agreeing if $|t_1 - t_2| \le \delta$ (we set $\delta=0.5$ seconds); if at least two out of three annotators satisfy this constraint, we adopt one of the agreed timestamps.
We then fine-tune an event-aware Qwen3-VL-32B model~\citep{Qwen3-VL} on the manually labeled subset and apply it to the remaining 47.1\% videos for large-scale filtering and boundary prediction, followed by manual verification to ensure logical and temporal consistency.
To characterize annotation consistency, we report the human triage distribution and boundary-agreement protocol under the same $\delta=0.5$s matching rule in Appendix~\ref{sup:guide}.

\subsection{Comprehensive Annotation}
\label{mev_step_4}
MEV adopts a comprehensive annotation scheme with \textbf{global} (video-level structured captions) and \textbf{event} (segments with captions, camera descriptions, and semantic attributes).

\noindent \textbf{Global structured captions.}
For each long-take video, we collect a structured set of global fields that summarize the overall context and dominant properties across the entire video: background, theme, style, shot type, camera movement, lighting, atmosphere, and subject profile.
These fields are designed to be composable: they can be concatenated into a global detailed description, or selectively sampled as conditioning signals for controllable generation.

\noindent \textbf{Multi-granularity captions.}
We provide three caption granularities---short, middle, and long---with lengths of around 20/80/200 words to balance prompt compactness and descriptive richness.
Short captions focus on the main presence. Middle and long captions both explicitly draw on the structured fields (\eg, lighting, style, subject profile); long captions integrate them more extensively and add finer procedural details to support fine-grained text-to-video generation.

\noindent \textbf{Event-level camera-aware annotation.}
Camera dynamics are not exclusive to the global level: even within a long take, filming motion can change across event segments.
Following the camera-motion taxonomy studied in~\citep{lin2025towards}, we annotate each event with a set of labels from 14 common motion types (dolly-in/out, pedestal-up/down, truck-left/right, pan-left/right, tilt-up/down, zoom-in/out, roll-CW/CCW) plus static, accompanied by a short natural-language description to explain the specific movement in that segment.
We also describe focus shifts in natural language, specifying the subjects or regions that become sharp or blurred as optical focus changes.
This provides an explicit signal for learning temporally evolving camera behavior and for conditioning camera-controllable generation.

\noindent \textbf{Event-level semantic attributes.}
Beyond event captions, we provide non-exclusive binary attributes describing event changes (\eg, appearance and disappearance), participant composition (\eg, multiple people, animals, and children), and activity or scene context (\eg, cooking, sports, and travel).
These extensible fields support subset retrieval and targeted training, with detailed actions described by the event captions.

\noindent \textbf{Video quality scores.}
To quantify video quality at both global and event levels, we compute a set of VBench metrics~\citep{huang2024vbench,zheng2025vbench}, including temporal flickering, subject consistency, background consistency, aesthetic quality, imaging quality, and motion smoothness.
These scores serve as both references for dataset analysis and practical knobs for curating subsets with different quality profiles.

We use Gemini 2.5 Pro~\citep{Comanicietal2025} to produce global structured captions and event-level annotations, motivated by its stronger action-centric wording for event captions and its improved sensitivity to camera motion compared with Qwen3-VL (22\% vs.\ 10\% exact-set accuracy in our manual evaluation), which frequently predicts ``static'' for moving-camera clips.
Concretely, we prompt the model with the full long-take video to obtain global fields, and with each segmented event clip to obtain event captions, camera-motion descriptions, and semantic attributes, while enforcing a JSON schema for scalable post-processing.
Meanwhile, Gemini can still confuse certain motion labels (\eg, mixing truck-left with pan-left); we therefore include a dedicated motion description and perform manual verification to correct obvious inconsistencies.
We further audit VLM-generated annotations against human labels and summarize their reliability across camera-motion labels and semantic attributes, with detailed accuracy results and representative error patterns provided in Appendix~\ref{sup:stats}.

\subsection{Statistical Comparison and Analysis}
\label{mev_anal}
\begin{figure*}[t]
  \centering
   \includegraphics[width=0.95\linewidth]{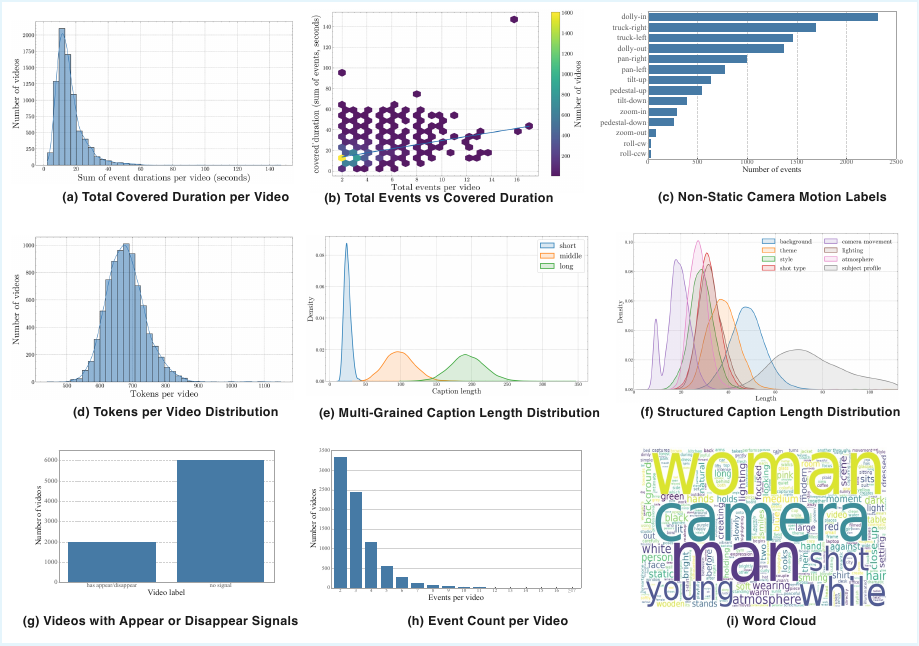}
   \caption{Statistical distributions of our \textbf{MEV} from different perspectives.}
   \label{fig:stats}
   \vspace{-1.5em}
\end{figure*}

\noindent \textbf{Comparison with popular video-text datasets.}
Tab.~\ref{tab:dataset} compares MEV with representative video--text datasets for text-to-video generation. Rather than matching web-scale caption-only corpora in raw size, MEV focuses on dense temporally grounded supervision per video. MEV stands out by jointly providing (i) high-quality long-take videos (720p--4K), (ii) explicit temporal grounding, and (iii) a comprehensive global--event annotation suite, including multi-granularity structured captions, camera-perspective descriptions, semantic attributes, and generation-relevant quality scores. 

\noindent \textbf{Statistics from multiple perspectives.}
Fig.~\ref{fig:stats} summarizes MEV statistics. The sum of covered durations per video concentrates around tens of seconds with a long tail (a). (b) relates event count to covered duration for videos with 2--17 events, (c) shows the distribution of non-static camera-motion labels, (d) depicts the distribution of total caption tokens per video. As shown in (e), the three granularities of global captions exhibit clear length separation, complemented by the length analysis for various structured fields in (f). Furthermore, (g) analyzes representative semantic attributes—using appearance/disappearance as an example—and reports their counts, (h) shows event count per video, and (i) provides an intuitive word cloud to visualize the captions.
We also provide additional camera-motion statistics and analyses of caption granularity and video--text retrieval in Appendix~\ref{sup:stats}.

\section{Experiments}
\label{sec:ex}
\begin{figure*}[t]
  \centering
   \includegraphics[width=1.0\linewidth]{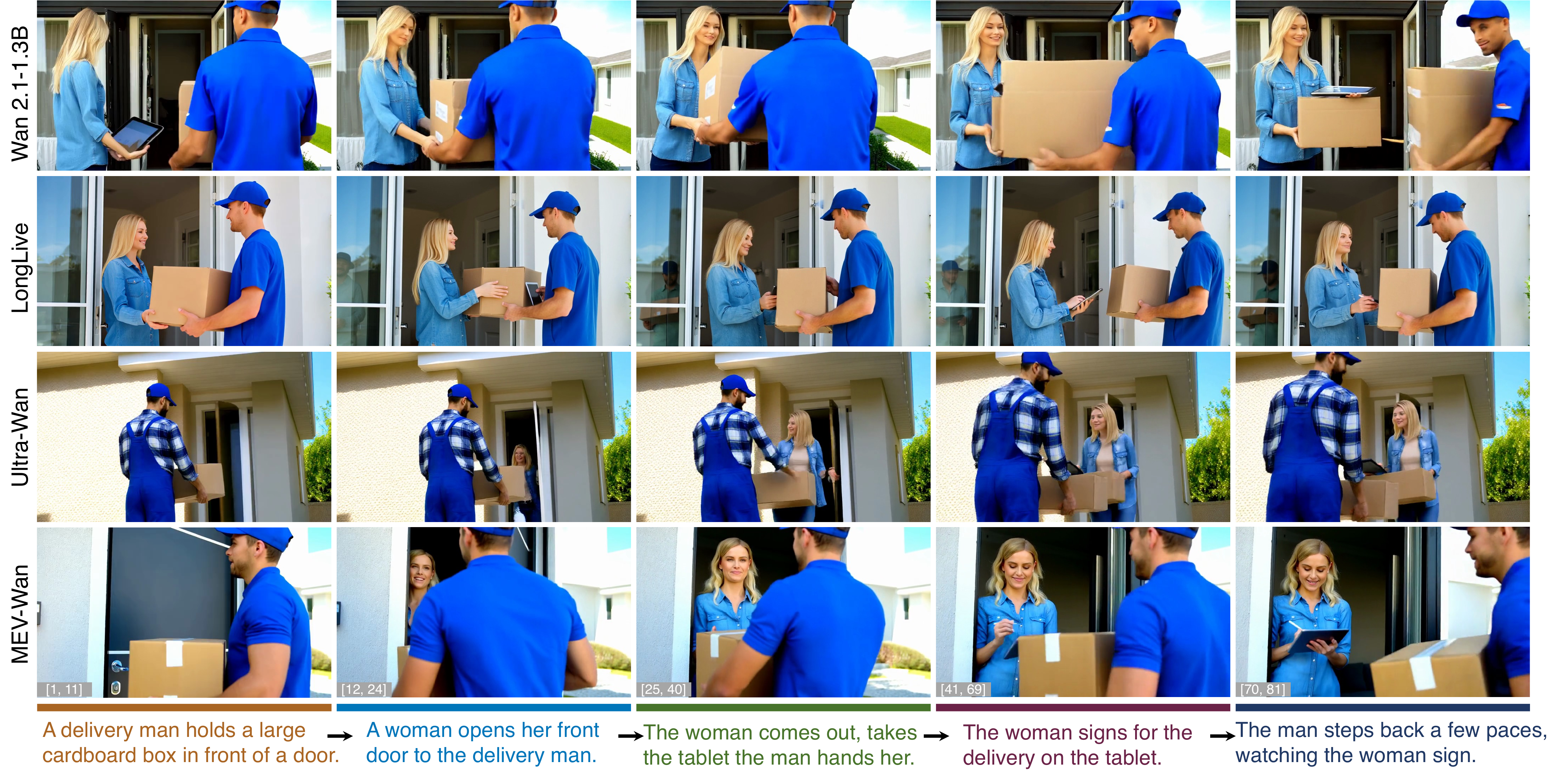}
   \vspace{-0.5em}
   \caption{\textbf{Qualitative results of T2V.} Wan2.1 struggles from the first event onward, exhibiting disordered temporal event structure and poor utilization of event information. LongLive further suffers from severe drifting (\eg, a human reflection in the glass introduced in the second key frame persists in subsequent frames). Ultra-Wan captures the event structure better, but shows increasing misalignment after the third event (\eg, the box splits into two) and exhibits video stagnation. MEV-Wan is the only method that generates all events with smooth transitions and consistent content. }
   \label{fig:obs}
   \vspace{-0.5em}
\end{figure*}

\subsection{Experimental Settings}

\noindent \textbf{Multi-event adaptation on Wan.}
Inspired by MinT~\citep{wu2025mind}, we adapt Wan for multi-event video generation using three coordinated components. The global cross-attention branch keeps whole-video semantics and style coherence from the global prompt, while the temporal cross-attention branch enforces event-level conditioning so that event captions are grounded to their corresponding temporal regions. ReRoPE provides the temporal positional bias that links event text tokens to frame-time assignments, enabling ordered event control with smooth transitions rather than hard boundaries. Training samples are built with contiguous event grouping and duration-aware frame allocation, so temporal supervision is consistent with event order in the data. In this work, we finetune with LoRA~\citep{hu2022lora} plugins to enable efficient adaptation. We refer to this approach—training Wan on the MEV dataset for multi-event video generation—as \textbf{MEV-Wan}. Architectural and training details are provided in Appendix~\ref{sup:mevg}.

\begin{wraptable}[12]{r}{0.48\textwidth}
\centering
\caption{\textbf{MEV-Eval event-centric generation results.} Metrics are averaged over the same five random 128-prompt draws as Table~\ref{tab:metrics}.}
\label{tab:mev_eval}
\setlength{\tabcolsep}{3pt}
\renewcommand{\arraystretch}{1.08}
\begin{adjustbox}{max width=\linewidth}
\begin{tabular}{lcccc}
\toprule
Method &
\makecell{EEA $\uparrow$} &
\makecell{EOA $\uparrow$} &
\makecell{TGA $\uparrow$} &
\makecell{BTC $\uparrow$} \\
\midrule
Wan2.1-1.3B & 0.42 & 0.20 & 0.33 & 0.41 \\
LongLive    & 0.47 & 0.39 & 0.46 & 0.58 \\
Ultra-Wan   & 0.46 & 0.33 & 0.42 & 0.50 \\
MEV-Wan     & \textbf{0.61} & \textbf{0.45} & \textbf{0.60} & \textbf{0.64} \\
\bottomrule
\end{tabular}
\end{adjustbox}
\end{wraptable}

\noindent \textbf{Baselines.}
All four methods derive from Wan2.1-1.3B. We evaluate the original Wan model and the official LongLive~\citep{yang2025longlive} checkpoint without further fine-tuning; LongLive uses autoregressive generation with prompt switching. Ultra-Wan and MEV-Wan use the same Wan backbone and MinT-style multi-event adaptation. They are fine-tuned on an UltraVideo-Long~\citep{xue2025ultravideo} subset and MEV, respectively, with the same number of videos. Ultra-Wan uses caption-inferred events with uniform temporal boundaries, whereas MEV-Wan uses verified events and grounded boundaries. Their comparison evaluates MEV as a complete training resource, including video content, captions, and temporal annotations. Method configurations and Ultra-Wan data construction are detailed in Appendix~\ref{sup:mevg}.

\noindent \textbf{Implementation details.}
We conduct generation experiments using Wan2.1-T2V-1.3B. We split MEV 80/20 at the original-video level before constructing contiguous groups of 2--5 events. All event groups derived from the same original video remain in the same split (Appendix~\ref{sup:mevg}). For training, all video data is processed at a spatial resolution of $480 \times 832$ with a frame count of 81. The model is conditioned on both the global middle caption and the corresponding event captions. We utilize a ReRoPE rescale length of 8 and a learning rate of $1 \times 10^{-5}$.

\subsection{MEV-Eval: Event-Centric Evaluation Protocol.}

\noindent \textbf{MEV-Eval.}
VBench metrics are useful for measuring general video quality, but they do not directly evaluate whether a generated video executes a specified sequence of events. We therefore introduce MEV-Eval, an event-centric evaluation protocol for long-take multi-event generation. It targets four failure modes: omitted or weakly realized events, incorrect event order, poor temporal localization, and incoherent transitions near event boundaries. Event Execution Accuracy (EEA) measures whether each event is visually executed; Event Order Accuracy (EOA) evaluates pairwise ordering; Temporal Grounding Accuracy (TGA) checks alignment with the assigned temporal region; and Boundary Transition Coherence (BTC) evaluates visual and semantic continuity across adjacent events. We compute all four metrics with VLM-assisted judging, using the same event list and duration assignment for all compared methods. As shown in Table~\ref{tab:mev_eval}, MEV-Wan obtains the strongest event-centric scores. On a fixed subset of 128 test prompts, human and Gemini evaluations both rank MEV-Wan first on all four metrics. Full metric definitions, the Gemini 2.5 Pro judging setup, and the human-validation metrics are provided in Appendix~\ref{sup:mev_eval}.

\subsection{MEV Benefits Multi-Event Video Generation.}

The experimental results demonstrate that the MEV dataset enhances the performance of models in the multi-event setting. Thanks to the structured captions and duration-aware temporal supervision inherent in our dataset, MEV-Wan effectively grounds event-level information, resulting in superior temporal coherence and content consistency. 
As illustrated in Figure~\ref{fig:obs}, baselines such as Wan2.1 and LongLive suffer from disordered event structures or severe content drifting. Ultra-Wan preserves event structure to some extent, but it suffers from event–text misalignment and video stagnation, suggesting that clearer grounding in the data is more conducive to effective temporal supervision. 
\begin{table}[t]
\centering
\caption{\textbf{Quantitative T2V Results.} We evaluate on 128 samples from the MEV test set and report the average over five random draws. Metrics include the six VBench dimensions.}
\label{tab:metrics}
\setlength{\tabcolsep}{3pt}
\renewcommand{\arraystretch}{1.15}
\begin{adjustbox}{max width=\linewidth}
\begin{tabular}{lcccccc}
\toprule
Method &
\makecell{Subject\\Consistency $\uparrow$} &
\makecell{Background\\Consistency $\uparrow$} &
\makecell{Temporal\\Flickering $\uparrow$} &
\makecell{Motion\\Smoothness $\uparrow$} &
\makecell{Aesthetic\\Quality $\uparrow$} &
\makecell{Imaging\\Quality $\uparrow$} \\
\midrule
Wan2.1-1.3B & 0.954 & 0.949 & 0.989 & 0.991 & 0.585 & 61.674 \\
LongLive    & 0.949 & 0.950 & 0.986 & 0.990 & 0.504 & 61.953 \\
Ultra-Wan   & 0.936 & 0.956 & 0.990 & 0.987 & 0.557 & 63.328 \\
MEV-Wan     & 0.960 & 0.959 & 0.990 & 0.991 & 0.590 & 64.924 \\
\bottomrule
\end{tabular}
\end{adjustbox}
\vspace{-0.5em}
\end{table}
Across generic visual metrics, MEV-Wan maintains the fidelity and smoothness of the Wan backbone while improving imaging quality, indicating that event-grounded adaptation does not trade temporal control for visual quality. The event-centric comparison further highlights the advantage of MEV-Wan in executing events in order, aligning them with intended temporal windows, and reducing transition drift. Notably, Ultra-Wan remains competitive on VBench but lags on MEV-Eval, suggesting the value of MEV's curated videos and temporally grounded event annotations for multi-event generation.
These findings suggest that fine-grained temporal alignment in MEV helps maintain coherent event progression and stable visual quality across multi-stage narratives.

\Needspace{12\baselineskip}
\section{Conclusion}
We present MEV, a long-take multi-event text--video dataset featuring precise temporal grounding and comprehensive global-event captions. MEV is constructed via a scalable pipeline featuring rule-based atomic grounding for event segmentation and a global-event annotation scheme with multi-granularity structured captions, camera-aware descriptions, and semantic attributes. We provide annotation-quality analyses and an event-centric evaluation protocol to better assess whether generated videos execute ordered events with coherent temporal transitions. Finally, we show that MEV serves as effective supervision for multi-event generation by adapting Wan with temporally grounded event conditioning, leading to improved long-take event transitions and more coherent content.

\clearpage
\flushbottom

\begin{ack}
We would like to acknowledge the support from NSF Award No.~2229881, AI Institute for Societal Decision Making (AI-SDM), the National Institutes of Health (NIH) under Contract R01HL159805, and grants from Quris AI, Florin Court Capital, MBZUAI-WIS Joint Program, and the Al Deira Causal Education project.
\end{ack}

\bibliographystyle{unsrtnat}
\bibliography{references}

\begin{thebibliography}{70}
\providecommand{\natexlab}[1]{#1}
\providecommand{\url}[1]{\texttt{#1}}
\expandafter\ifx\csname urlstyle\endcsname\relax
  \providecommand{\doi}[1]{doi: #1}\else
  \providecommand{\doi}{doi: \begingroup \urlstyle{rm}\Url}\fi

\bibitem[Whitehead(1920)]{whitehead1920concept}
Alfred~North Whitehead.
\newblock \emph{The Concept of Nature}.
\newblock Cambridge University Press, 1920.

\bibitem[Zacks et~al.(2007)Zacks, Speer, Swallow, Braver, and
  Reynolds]{zacks2007event}
Jeffrey~M Zacks, Nicole~K Speer, Khena~M Swallow, Todd~S Braver, and Jeremy~R
  Reynolds.
\newblock Event perception: a mind-brain perspective.
\newblock \emph{Psychological bulletin}, 133\penalty0 (2):\penalty0 273--293,
  2007.

\bibitem[James(1890)]{james1890principles}
William James.
\newblock \emph{The principles of psychology}, volume~1.
\newblock Macmillan London, 1890.

\bibitem[Cai et~al.(2025{\natexlab{a}})Cai, Yang, Zhang, Guo, Xiao, Yang, Xu,
  Yang, Yuille, Guibas, et~al.]{cai2025mixture}
Shengqu Cai, Ceyuan Yang, Lvmin Zhang, Yuwei Guo, Junfei Xiao, Ziyan Yang,
  Yinghao Xu, Zhenheng Yang, Alan Yuille, Leonidas Guibas, et~al.
\newblock Mixture of contexts for long video generation.
\newblock \emph{arXiv preprint arXiv:2508.21058}, 2025{\natexlab{a}}.

\bibitem[Guo et~al.(2025)Guo, Yang, Yang, Ma, Lin, Yang, Lin, and
  Jiang]{guo2025long}
Yuwei Guo, Ceyuan Yang, Ziyan Yang, Zhibei Ma, Zhijie Lin, Zhenheng Yang, Dahua
  Lin, and Lu~Jiang.
\newblock Long context tuning for video generation.
\newblock \emph{arXiv preprint arXiv:2503.10589}, 2025.

\bibitem[Wang et~al.(2025{\natexlab{a}})Wang, Shi, Li, Bian, Liu, Lu, Wang,
  Wan, Gai, and Jia]{wang2025multishotmaster}
Qinghe Wang, Xiaoyu Shi, Baolu Li, Weikang Bian, Quande Liu, Huchuan Lu, Xintao
  Wang, Pengfei Wan, Kun Gai, and Xu~Jia.
\newblock Multishotmaster: A controllable multi-shot video generation
  framework.
\newblock \emph{arXiv preprint arXiv:2512.03041}, 2025{\natexlab{a}}.

\bibitem[Henschel et~al.(2025)Henschel, Khachatryan, Poghosyan, Hayrapetyan,
  Tadevosyan, Wang, Navasardyan, and Shi]{henschel2025streamingt2v}
Roberto Henschel, Levon Khachatryan, Hayk Poghosyan, Daniil Hayrapetyan, Vahram
  Tadevosyan, Zhangyang Wang, Shant Navasardyan, and Humphrey Shi.
\newblock Streamingt2v: Consistent, dynamic, and extendable long video
  generation from text.
\newblock In \emph{Proceedings of the Computer Vision and Pattern Recognition
  Conference}, pages 2568--2577, 2025.

\bibitem[Cai et~al.(2025{\natexlab{b}})Cai, Cun, Li, Liu, Zhang, Zhang, Shan,
  and Yue]{cai2025ditctrl}
Minghong Cai, Xiaodong Cun, Xiaoyu Li, Wenze Liu, Zhaoyang Zhang, Yong Zhang,
  Ying Shan, and Xiangyu Yue.
\newblock Ditctrl: Exploring attention control in multi-modal diffusion
  transformer for tuning-free multi-prompt longer video generation.
\newblock In \emph{Proceedings of the Computer Vision and Pattern Recognition
  Conference}, pages 7763--7772, 2025{\natexlab{b}}.

\bibitem[Yang et~al.(2025{\natexlab{a}})Yang, Huang, Chu, Xiao, Zhao, Wang, Li,
  Xie, Chen, Lu, et~al.]{yang2025longlive}
Shuai Yang, Wei Huang, Ruihang Chu, Yicheng Xiao, Yuyang Zhao, Xianbang Wang,
  Muyang Li, Enze Xie, Yingcong Chen, Yao Lu, et~al.
\newblock Longlive: Real-time interactive long video generation.
\newblock \emph{arXiv preprint arXiv:2509.22622}, 2025{\natexlab{a}}.

\bibitem[Lu et~al.(2025)Lu, Zeng, Li, Ouyang, Wang, Cheng, Zhu, Cao, Zhang,
  Zhu, et~al.]{lu2025reward}
Yunhong Lu, Yanhong Zeng, Haobo Li, Hao Ouyang, Qiuyu Wang, Ka~Leong Cheng,
  Jiapeng Zhu, Hengyuan Cao, Zhipeng Zhang, Xing Zhu, et~al.
\newblock Reward forcing: Efficient streaming video generation with rewarded
  distribution matching distillation.
\newblock \emph{arXiv preprint arXiv:2512.04678}, 2025.

\bibitem[Po et~al.(2025)Po, Chan, Chen, and Wetzstein]{po2025bagger}
Ryan Po, Eric~Ryan Chan, Changan Chen, and Gordon Wetzstein.
\newblock Bagger: Backwards aggregation for mitigating drift in autoregressive
  video diffusion models.
\newblock \emph{arXiv preprint arXiv:2512.12080}, 2025.

\bibitem[Chen et~al.(2024{\natexlab{a}})Chen, Siarohin, Menapace, Deyneka,
  Chao, Jeon, Fang, Lee, Ren, Yang, et~al.]{chen2024panda}
Tsai-Shien Chen, Aliaksandr Siarohin, Willi Menapace, Ekaterina Deyneka,
  Hsiang-wei Chao, Byung~Eun Jeon, Yuwei Fang, Hsin-Ying Lee, Jian Ren,
  Ming-Hsuan Yang, et~al.
\newblock Panda-70m: Captioning 70m videos with multiple cross-modality
  teachers.
\newblock In \emph{Proceedings of the IEEE/CVF Conference on Computer Vision
  and Pattern Recognition}, pages 13320--13331, 2024{\natexlab{a}}.

\bibitem[Tan et~al.(2024)Tan, Yang, Qin, and Li]{tan2024vidgen}
Zhiyu Tan, Xiaomeng Yang, Luozheng Qin, and Hao Li.
\newblock Vidgen-1m: A large-scale dataset for text-to-video generation.
\newblock \emph{arXiv preprint arXiv:2408.02629}, 2024.

\bibitem[Nan et~al.(2024)Nan, Xie, Zhou, Fan, Yang, Chen, Li, Yang, and
  Tai]{nan2024openvid}
Kepan Nan, Rui Xie, Penghao Zhou, Tiehan Fan, Zhenheng Yang, Zhijie Chen, Xiang
  Li, Jian Yang, and Ying Tai.
\newblock Openvid-1m: A large-scale high-quality dataset for text-to-video
  generation.
\newblock \emph{arXiv preprint arXiv:2407.02371}, 2024.

\bibitem[Wang et~al.(2025{\natexlab{b}})Wang, Shi, Ou, Chen, Lin, Wang, Jiang,
  Yang, Zheng, Tao, et~al.]{wang2025koala}
Qiuheng Wang, Yukai Shi, Jiarong Ou, Rui Chen, Ke~Lin, Jiahao Wang, Boyuan
  Jiang, Haotian Yang, Mingwu Zheng, Xin Tao, et~al.
\newblock Koala-36m: A large-scale video dataset improving consistency between
  fine-grained conditions and video content.
\newblock In \emph{Proceedings of the Computer Vision and Pattern Recognition
  Conference}, pages 8428--8437, 2025{\natexlab{b}}.

\bibitem[Bolya et~al.(2025)Bolya, Huang, Sun, Cho, Madotto, Wei, Ma, Zhi,
  Rajasegaran, Rasheed, et~al.]{bolya2025perception}
Daniel Bolya, Po-Yao Huang, Peize Sun, Jang~Hyun Cho, Andrea Madotto, Chen Wei,
  Tengyu Ma, Jiale Zhi, Jathushan Rajasegaran, Hanoona Rasheed, et~al.
\newblock Perception encoder: The best visual embeddings are not at the output
  of the network.
\newblock \emph{arXiv preprint arXiv:2504.13181}, 2025.

\bibitem[Heilbron et~al.(2015)Heilbron, Escorcia, Ghanem, and
  Niebles]{Heilbron2015ActivityNetAL}
Fabian~Caba Heilbron, Victor Escorcia, Bernard Ghanem, and Juan~Carlos Niebles.
\newblock Activitynet: A large-scale video benchmark for human activity
  understanding.
\newblock \emph{2015 IEEE Conference on Computer Vision and Pattern Recognition
  (CVPR)}, pages 961--970, 2015.
\newblock URL \url{https://api.semanticscholar.org/CorpusID:1710722}.

\bibitem[Oncescu et~al.(2021)Oncescu, Henriques, Liu, Zisserman, and
  Albanie]{oncescu2021queryd}
Andreea-Maria Oncescu, Joao~F Henriques, Yang Liu, Andrew Zisserman, and Samuel
  Albanie.
\newblock Queryd: A video dataset with high-quality text and audio narrations.
\newblock In \emph{ICASSP 2021-2021 IEEE International Conference on Acoustics,
  Speech and Signal Processing (ICASSP)}, pages 2265--2269. IEEE, 2021.

\bibitem[Soldan et~al.(2022)Soldan, Pardo, Alc{\'a}zar, Caba, Zhao, Giancola,
  and Ghanem]{soldan2022mad}
Mattia Soldan, Alejandro Pardo, Juan~Le{\'o}n Alc{\'a}zar, Fabian Caba, Chen
  Zhao, Silvio Giancola, and Bernard Ghanem.
\newblock Mad: A scalable dataset for language grounding in videos from movie
  audio descriptions.
\newblock In \emph{Proceedings of the IEEE/CVF Conference on Computer Vision
  and Pattern Recognition}, pages 5026--5035, 2022.

\bibitem[Chen et~al.(2024{\natexlab{b}})Chen, Liu, He, Chen, Gan, Ma, Zhong,
  Zhang, Wang, Lin, et~al.]{chen2024mecd}
Tieyuan Chen, Huabin Liu, Tianyao He, Yihang Chen, Chaofan Gan, Xiao Ma, Cheng
  Zhong, Yang Zhang, Yingxue Wang, Hui Lin, et~al.
\newblock Mecd: Unlocking multi-event causal discovery in video reasoning.
\newblock \emph{Advances in Neural Information Processing Systems},
  37:\penalty0 92554--92580, 2024{\natexlab{b}}.

\bibitem[Huang et~al.(2020)Huang, Xiong, Rao, Wang, and Lin]{huang2020movienet}
Qingqiu Huang, Yu~Xiong, Anyi Rao, Jiaze Wang, and Dahua Lin.
\newblock Movienet: A holistic dataset for movie understanding.
\newblock In \emph{European conference on computer vision}, pages 709--727.
  Springer, 2020.

\bibitem[Wang and Yang(2025)]{wang2025videoufo}
Wenhao Wang and Yi~Yang.
\newblock Videoufo: A million-scale user-focused dataset for text-to-video
  generation.
\newblock \emph{arXiv preprint arXiv:2503.01739}, 2025.

\bibitem[Wan et~al.(2025)Wan, Wang, Ai, Wen, Mao, Xie, Chen, Yu, Zhao, Yang,
  et~al.]{wan2025wan}
Team Wan, Ang Wang, Baole Ai, Bin Wen, Chaojie Mao, Chen-Wei Xie, Di~Chen,
  Feiwu Yu, Haiming Zhao, Jianxiao Yang, et~al.
\newblock Wan: Open and advanced large-scale video generative models.
\newblock \emph{arXiv preprint arXiv:2503.20314}, 2025.

\bibitem[Miech et~al.(2019)Miech, Zhukov, Alayrac, Tapaswi, Laptev, and
  Sivic]{miech2019howto100m}
Antoine Miech, Dimitri Zhukov, Jean-Baptiste Alayrac, Makarand Tapaswi, Ivan
  Laptev, and Josef Sivic.
\newblock Howto100m: Learning a text-video embedding by watching hundred
  million narrated video clips.
\newblock In \emph{Proceedings of the IEEE/CVF international conference on
  computer vision}, pages 2630--2640, 2019.

\bibitem[Bain et~al.(2021)Bain, Nagrani, Varol, and Zisserman]{bain2021frozen}
Max Bain, Arsha Nagrani, G{\"u}l Varol, and Andrew Zisserman.
\newblock Frozen in time: A joint video and image encoder for end-to-end
  retrieval.
\newblock In \emph{Proceedings of the IEEE/CVF international conference on
  computer vision}, pages 1728--1738, 2021.

\bibitem[Xue et~al.(2022)Xue, Hang, Zeng, Sun, Liu, Yang, Fu, and
  Guo]{xue2022advancing}
Hongwei Xue, Tiankai Hang, Yanhong Zeng, Yuchong Sun, Bei Liu, Huan Yang,
  Jianlong Fu, and Baining Guo.
\newblock Advancing high-resolution video-language representation with
  large-scale video transcriptions.
\newblock In \emph{Proceedings of the IEEE/CVF Conference on Computer Vision
  and Pattern Recognition}, pages 5036--5045, 2022.

\bibitem[Wang et~al.(2023)Wang, He, Li, Li, Yu, Ma, Li, Chen, Chen, Wang,
  et~al.]{wang2023internvid}
Yi~Wang, Yinan He, Yizhuo Li, Kunchang Li, Jiashuo Yu, Xin Ma, Xinhao Li, Guo
  Chen, Xinyuan Chen, Yaohui Wang, et~al.
\newblock Internvid: A large-scale video-text dataset for multimodal
  understanding and generation.
\newblock \emph{arXiv preprint arXiv:2307.06942}, 2023.

\bibitem[Ju et~al.(2024)Ju, Gao, Zhang, Yuan, Wang, Zeng, Xiong, Xu, and
  Shan]{ju2024miradata}
Xuan Ju, Yiming Gao, Zhaoyang Zhang, Ziyang Yuan, Xintao Wang, Ailing Zeng,
  Yu~Xiong, Qiang Xu, and Ying Shan.
\newblock Miradata: A large-scale video dataset with long durations and
  structured captions.
\newblock \emph{Advances in Neural Information Processing Systems},
  37:\penalty0 48955--48970, 2024.

\bibitem[Xiong et~al.(2024)Xiong, Wang, Zhou, Lin, Feng, and Liu]{xiong2024lvd}
Tianwei Xiong, Yuqing Wang, Daquan Zhou, Zhijie Lin, Jiashi Feng, and Xihui
  Liu.
\newblock Lvd-2m: A long-take video dataset with temporally dense captions.
\newblock \emph{Advances in Neural Information Processing Systems},
  37:\penalty0 16623--16644, 2024.

\bibitem[Lin et~al.(2024)Lin, Ge, Cheng, Li, Zhu, Wang, He, Ye, Yuan, Chen,
  et~al.]{lin2024open}
Bin Lin, Yunyang Ge, Xinhua Cheng, Zongjian Li, Bin Zhu, Shaodong Wang, Xianyi
  He, Yang Ye, Shenghai Yuan, Liuhan Chen, et~al.
\newblock Open-sora plan: Open-source large video generation model.
\newblock \emph{arXiv preprint arXiv:2412.00131}, 2024.

\bibitem[Xue et~al.(2025)Xue, Zhang, Hu, He, Chen, Cai, Wang, Wang, Liu, Li,
  et~al.]{xue2025ultravideo}
Zhucun Xue, Jiangning Zhang, Teng Hu, Haoyang He, Yinan Chen, Yuxuan Cai,
  Yabiao Wang, Chengjie Wang, Yong Liu, Xiangtai Li, et~al.
\newblock Ultravideo: High-quality uhd video dataset with comprehensive
  captions.
\newblock \emph{arXiv preprint arXiv:2506.13691}, 2025.

\bibitem[Yang et~al.(2025{\natexlab{b}})Yang, Ren, Yuille, and
  Wang]{yang2025vimix}
Timing Yang, Sucheng Ren, Alan Yuille, and Feng Wang.
\newblock Vimix-14m: A curated multi-source video-text dataset with long-form,
  high-quality captions and crawl-free access.
\newblock \emph{arXiv preprint arXiv:2511.18382}, 2025{\natexlab{b}}.

\bibitem[{Mixkit}(n.d.)]{mixkit_website}
{Mixkit}.
\newblock Mixkit: Free stock video \& music library.
\newblock \url{https://mixkit.co/}, n.d.
\newblock Accessed: 2026-03-03.

\bibitem[Soucek and Lokoc(2024)]{soucek2024transnet}
Tom{\'a}s Soucek and Jakub Lokoc.
\newblock Transnet v2: An effective deep network architecture for fast shot
  transition detection.
\newblock In \emph{Proceedings of the 32nd ACM International Conference on
  Multimedia}, pages 11218--11221, 2024.

\bibitem[Teed and Deng(2020)]{teed2020raft}
Zachary Teed and Jia Deng.
\newblock Raft: Recurrent all-pairs field transforms for optical flow.
\newblock In \emph{European conference on computer vision}, pages 402--419.
  Springer, 2020.

\bibitem[Bai et~al.(2025)Bai, Cai, Chen, Chen, Chen, Cheng, Deng, Ding, Gao,
  Ge, Ge, Guo, Huang, Huang, Huang, Hui, Jiang, Li, Li, Li, Li, Lin, Lin, Liu,
  Liu, Liu, Liu, Liu, Liu, Lu, Luo, Lv, Men, Meng, Ren, Ren, Song, Sun, Tang,
  Tu, Wan, Wang, Wang, Wang, Wang, Xie, Xu, Xu, Xu, Yang, Yang, Yang, Yang, Yu,
  Zhang, Zhang, Zhang, Zheng, Zhong, Zhou, Zhou, Zhou, Zhu, and Zhu]{Qwen3-VL}
Shuai Bai, Yuxuan Cai, Ruizhe Chen, Keqin Chen, Xionghui Chen, Zesen Cheng,
  Lianghao Deng, Wei Ding, Chang Gao, Chunjiang Ge, Wenbin Ge, Zhifang Guo,
  Qidong Huang, Jie Huang, Fei Huang, Binyuan Hui, Shutong Jiang, Zhaohai Li,
  Mingsheng Li, Mei Li, Kaixin Li, Zicheng Lin, Junyang Lin, Xuejing Liu,
  Jiawei Liu, Chenglong Liu, Yang Liu, Dayiheng Liu, Shixuan Liu, Dunjie Lu,
  Ruilin Luo, Chenxu Lv, Rui Men, Lingchen Meng, Xuancheng Ren, Xingzhang Ren,
  Sibo Song, Yuchong Sun, Jun Tang, Jianhong Tu, Jianqiang Wan, Peng Wang,
  Pengfei Wang, Qiuyue Wang, Yuxuan Wang, Tianbao Xie, Yiheng Xu, Haiyang Xu,
  Jin Xu, Zhibo Yang, Mingkun Yang, Jianxin Yang, An~Yang, Bowen Yu, Fei Zhang,
  Hang Zhang, Xi~Zhang, Bo~Zheng, Humen Zhong, Jingren Zhou, Fan Zhou, Jing
  Zhou, Yuanzhi Zhu, and Ke~Zhu.
\newblock Qwen3-vl technical report.
\newblock \emph{arXiv preprint arXiv:2511.21631}, 2025.

\bibitem[Comanici et~al.(2025)Comanici, Bieber, Schaekermann, Pasupat,
  Sachdeva, Dhillon, Blistein, Ram, Zhang, Rosen, and {Gemini Team,
  Google}]{Comanicietal2025}
Gheorghe Comanici, Eric Bieber, Mike Schaekermann, Ice Pasupat, Noveen
  Sachdeva, Inderjit Dhillon, Marcel Blistein, Ori Ram, Dan Zhang, Evan Rosen,
  and {Gemini Team, Google}.
\newblock Gemini 2.5: Pushing the frontier with advanced reasoning,
  multimodality, long context, and next generation agentic capabilities.
\newblock \emph{arXiv preprint arXiv:2507.06261}, 2025.
\newblock \doi{10.48550/arXiv.2507.06261}.

\bibitem[Lin et~al.(2025)Lin, Cen, Jiang, Karhade, Wang, Mitra, Ling, Huang,
  Liu, Chen, et~al.]{lin2025towards}
Zhiqiu Lin, Siyuan Cen, Daniel Jiang, Jay Karhade, Hewei Wang, Chancharik
  Mitra, Tiffany Ling, Yuhan Huang, Sifan Liu, Mingyu Chen, et~al.
\newblock Towards understanding camera motions in any video.
\newblock \emph{arXiv preprint arXiv:2504.15376}, 2025.

\bibitem[Huang et~al.(2024)Huang, He, Yu, Zhang, Si, Jiang, Zhang, Wu, Jin,
  Chanpaisit, et~al.]{huang2024vbench}
Ziqi Huang, Yinan He, Jiashuo Yu, Fan Zhang, Chenyang Si, Yuming Jiang, Yuanhan
  Zhang, Tianxing Wu, Qingyang Jin, Nattapol Chanpaisit, et~al.
\newblock Vbench: Comprehensive benchmark suite for video generative models.
\newblock In \emph{Proceedings of the IEEE/CVF Conference on Computer Vision
  and Pattern Recognition}, pages 21807--21818, 2024.

\bibitem[Zheng et~al.(2025)Zheng, Huang, Liu, Zou, He, Zhang, Gu, Zhang, He,
  Zheng, et~al.]{zheng2025vbench}
Dian Zheng, Ziqi Huang, Hongbo Liu, Kai Zou, Yinan He, Fan Zhang, Lulu Gu,
  Yuanhan Zhang, Jingwen He, Wei-Shi Zheng, et~al.
\newblock Vbench-2.0: Advancing video generation benchmark suite for intrinsic
  faithfulness.
\newblock \emph{arXiv preprint arXiv:2503.21755}, 2025.

\bibitem[Wu et~al.(2025)Wu, Siarohin, Menapace, Skorokhodov, Fang, Chordia,
  Gilitschenski, and Tulyakov]{wu2025mind}
Ziyi Wu, Aliaksandr Siarohin, Willi Menapace, Ivan Skorokhodov, Yuwei Fang,
  Varnith Chordia, Igor Gilitschenski, and Sergey Tulyakov.
\newblock Mind the time: Temporally-controlled multi-event video generation.
\newblock In \emph{Proceedings of the Computer Vision and Pattern Recognition
  Conference}, pages 23989--24000, 2025.

\bibitem[Hu et~al.(2022)Hu, Shen, Wallis, Allen-Zhu, Li, Wang, Wang, and
  Chen]{hu2022lora}
Edward~J Hu, Yelong Shen, Phillip Wallis, Zeyuan Allen-Zhu, Yuanzhi Li, Shean
  Wang, Lu~Wang, and Weizhu Chen.
\newblock {LoRA}: Low-rank adaptation of large language models.
\newblock In \emph{International Conference on Learning Representations}, 2022.
\newblock URL \url{https://openreview.net/forum?id=nZeVKeeFYf9}.

\bibitem[Soomro et~al.(2012)Soomro, Zamir, and Shah]{soomro2012ucf101}
Khurram Soomro, Amir~Roshan Zamir, and Mubarak Shah.
\newblock Ucf101: A dataset of 101 human actions classes from videos in the
  wild.
\newblock \emph{arXiv preprint arXiv:1212.0402}, 2012.

\bibitem[Kay et~al.(2017)Kay, Carreira, Simonyan, Zhang, Hillier,
  Vijayanarasimhan, Viola, Green, Back, Natsev, et~al.]{kay2017kinetics}
Will Kay, Joao Carreira, Karen Simonyan, Brian Zhang, Chloe Hillier, Sudheendra
  Vijayanarasimhan, Fabio Viola, Tim Green, Trevor Back, Paul Natsev, et~al.
\newblock The kinetics human action video dataset.
\newblock \emph{arXiv preprint arXiv:1705.06950}, 2017.

\bibitem[Tang et~al.(2019)Tang, Ding, Rao, Zheng, Zhang, Zhao, Lu, and
  Zhou]{tang2019coin}
Yansong Tang, Dajun Ding, Yongming Rao, Yu~Zheng, Danyang Zhang, Lili Zhao,
  Jiwen Lu, and Jie Zhou.
\newblock Coin: A large-scale dataset for comprehensive instructional video
  analysis.
\newblock In \emph{Proceedings of the IEEE/CVF Conference on Computer Vision
  and Pattern Recognition}, pages 1207--1216, 2019.

\bibitem[Sener et~al.(2022)Sener, Chatterjee, Shelepov, He, Singhania, Wang,
  and Yao]{sener2022assembly101}
Fadime Sener, Dibyadip Chatterjee, Daniel Shelepov, Kun He, Dipika Singhania,
  Robert Wang, and Angela Yao.
\newblock Assembly101: A large-scale multi-view video dataset for understanding
  procedural activities.
\newblock In \emph{Proceedings of the IEEE/CVF Conference on Computer Vision
  and Pattern Recognition}, pages 21096--21106, 2022.

\bibitem[Krishna et~al.(2017)Krishna, Hata, Ren, Fei-Fei, and
  Niebles]{krishna2017dense}
Ranjay Krishna, Kenji Hata, Frederic Ren, Li~Fei-Fei, and Juan~Carlos Niebles.
\newblock Dense-captioning events in videos.
\newblock In \emph{Proceedings of the IEEE International Conference on Computer
  Vision (ICCV)}, pages 706--715, 2017.

\bibitem[Mangalam et~al.(2023)Mangalam, Akshulakov, and
  Malik]{mangalam2023egoschema}
Karttikeya Mangalam, Raiymbek Akshulakov, and Jitendra Malik.
\newblock Egoschema: A diagnostic benchmark for very long-form video language
  understanding.
\newblock \emph{Advances in Neural Information Processing Systems},
  36:\penalty0 46212--46244, 2023.

\bibitem[OpenAI(2024)]{openai2024video}
OpenAI.
\newblock Video generation models as world simulators, 2024.

\bibitem[Li et~al.(2025)Li, Xu, Zhan, Mu, Li, Cheng, Chen, Chen, Ye, Wang,
  et~al.]{li2025openhumanvid}
Hui Li, Mingwang Xu, Yun Zhan, Shan Mu, Jiaye Li, Kaihui Cheng, Yuxuan Chen,
  Tan Chen, Mao Ye, Jingdong Wang, et~al.
\newblock Openhumanvid: A large-scale high-quality dataset for enhancing
  human-centric video generation.
\newblock In \emph{Proceedings of the Computer Vision and Pattern Recognition
  Conference}, pages 7752--7762, 2025.

\bibitem[Villegas et~al.(2022)Villegas, Babaeizadeh, Kindermans, Moraldo,
  Zhang, Saffar, Castro, Kunze, and Erhan]{villegas2022phenaki}
Ruben Villegas, Mohammad Babaeizadeh, Pieter-Jan Kindermans, Hernan Moraldo,
  Han Zhang, Mohammad~Taghi Saffar, Santiago Castro, Julius Kunze, and Dumitru
  Erhan.
\newblock Phenaki: Variable length video generation from open domain textual
  description.
\newblock \emph{arXiv preprint arXiv:2210.02399}, 2022.

\bibitem[Oh et~al.(2024)Oh, Jeong, Kim, Byeon, Kim, Kim, and Kim]{oh2024mevg}
Gyeongrok Oh, Jaehwan Jeong, Sieun Kim, Wonmin Byeon, Jinkyu Kim, Sungwoong
  Kim, and Sangpil Kim.
\newblock Mevg: Multi-event video generation with text-to-video models.
\newblock In \emph{European Conference on computer vision}, pages 401--418.
  Springer, 2024.

\bibitem[Yin et~al.(2024)Yin, Zhang, Zhang, Freeman, Durand, Shechtman, and
  Huang]{yin2024slow}
Tianwei Yin, Qiang Zhang, Richard Zhang, William~T Freeman, Fredo Durand, Eli
  Shechtman, and Xun Huang.
\newblock From slow bidirectional to fast causal video generators.
\newblock \emph{arXiv preprint arXiv:2412.07772}, 2024.

\bibitem[Huang et~al.(2025)Huang, Li, He, Zhou, and Shechtman]{huang2025self}
Xun Huang, Zhengqi Li, Guande He, Mingyuan Zhou, and Eli Shechtman.
\newblock Self forcing: Bridging the train-test gap in autoregressive video
  diffusion.
\newblock \emph{arXiv preprint arXiv:2506.08009}, 2025.

\bibitem[Kim et~al.(2025)Kim, Oh, Wang, Lee, and Shin]{kim2025tuning}
Subin Kim, Seoung~Wug Oh, Jui-Hsien Wang, Joon-Young Lee, and Jinwoo Shin.
\newblock Tuning-free multi-event long video generation via synchronized
  coupled sampling.
\newblock In \emph{Proceedings of the IEEE/CVF International Conference on
  Computer Vision (ICCV) Workshops}, pages 6477--6488, 2025.

\bibitem[Song et~al.(2020)Song, Meng, and Ermon]{song2020denoising}
Jiaming Song, Chenlin Meng, and Stefano Ermon.
\newblock Denoising diffusion implicit models.
\newblock \emph{arXiv preprint arXiv:2010.02502}, 2020.

\bibitem[Kim et~al.(2023)Kim, Lee, Choi, Jeong, Sohn, and
  Shin]{kim2023collaborative}
Subin Kim, Kyungmin Lee, June~Suk Choi, Jongheon Jeong, Kihyuk Sohn, and Jinwoo
  Shin.
\newblock Collaborative score distillation for consistent visual editing.
\newblock \emph{Advances in Neural Information Processing Systems},
  36:\penalty0 73232--73257, 2023.

\bibitem[Kara et~al.(2025)Kara, Singh, Liu, Ceylan, Rehg, and
  Hinz]{kara2025shotadapter}
Ozgur Kara, Krishna~Kumar Singh, Feng Liu, Duygu Ceylan, James~M Rehg, and
  Tobias Hinz.
\newblock Shotadapter: Text-to-multi-shot video generation with diffusion
  models.
\newblock In \emph{Proceedings of the Computer Vision and Pattern Recognition
  Conference}, pages 28405--28415, 2025.

\bibitem[Wang et~al.(2025{\natexlab{c}})Wang, Sheng, Cai, Zhang, Yan, Feng,
  Deng, and Ye]{wang2025echoshot}
Jiahao Wang, Hualian Sheng, Sijia Cai, Weizhan Zhang, Caixia Yan, Yachuang
  Feng, Bing Deng, and Jieping Ye.
\newblock Echoshot: Multi-shot portrait video generation.
\newblock In \emph{The Thirty-ninth Annual Conference on Neural Information
  Processing Systems}, 2025{\natexlab{c}}.

\bibitem[An et~al.(2025)An, Jia, Qiu, Zhou, Huang, Liu, Ren, Kahatapitiya, Liu,
  He, et~al.]{an2025onestory}
Zhaochong An, Menglin Jia, Haonan Qiu, Zijian Zhou, Xiaoke Huang, Zhiheng Liu,
  Weiming Ren, Kumara Kahatapitiya, Ding Liu, Sen He, et~al.
\newblock Onestory: Coherent multi-shot video generation with adaptive memory.
\newblock \emph{arXiv preprint arXiv:2512.07802}, 2025.

\bibitem[Luo et~al.(2022)Luo, Ji, Zhong, Chen, Lei, Duan, and
  Li]{luo2022clip4clip}
Huaishao Luo, Lei Ji, Ming Zhong, Yang Chen, Wen Lei, Nan Duan, and Tianrui Li.
\newblock Clip4clip: An empirical study of clip for end to end video clip
  retrieval and captioning.
\newblock \emph{Neurocomputing}, 508:\penalty0 293--304, 2022.

\bibitem[Ma et~al.(2022)Ma, Xu, Sun, Yan, Zhang, and Ji]{ma2022xclip}
Yiwei Ma, Guohai Xu, Xiaoshuai Sun, Ming Yan, Ji~Zhang, and Rongrong Ji.
\newblock {X-CLIP}: End-to-end multi-grained contrastive learning for
  video-text retrieval.
\newblock In \emph{Proceedings of the 30th ACM International Conference on
  Multimedia}, 2022.
\newblock \doi{10.1145/3503161.3547910}.

\bibitem[Zhu et~al.(2026)Zhu, Xie, Li, Shen, Deka, Shrivastava, Chen, and
  Zhang]{zhu2026mova}
Peiyuan Zhu, Shaoan Xie, Zijian Li, Yifan Shen, Namrata Deka, Harsh
  Shrivastava, Guangyi Chen, and Kun Zhang.
\newblock {MoVA}: Learning asymmetric dual projections for modular long
  video-text alignment.
\newblock In \emph{European Conference on Computer Vision (ECCV)}, pages
  373--393, 2026.
\newblock URL \url{https://arxiv.org/abs/2607.00858}.

\bibitem[{Kuaishou Technology}(2025)]{kling2025video26}
{Kuaishou Technology}.
\newblock {Kling AI} launches {Video 2.6} model with ``simultaneous
  audio-visual generation'' capability, redefining {AI} video creation
  workflow, December 2025.
\newblock URL
  \url{https://ir.kuaishou.com/news-releases/news-release-details/kling-ai-launches-video-26-model-simultaneous-audio-visual}.

\bibitem[Oquab et~al.(2023)Oquab, Darcet, Moutakanni, Vo, Szafraniec, Khalidov,
  Fernandez, Haziza, Massa, El-Nouby, Assran, Ballas, Galuba, Howes, Huang, Li,
  Misra, Rabbat, Sharma, Synnaeve, Xu, J{\'e}gou, Mairal, Labatut, Joulin, and
  Bojanowski]{oquab2023dinov2}
Maxime Oquab, Timoth{\'e}e Darcet, Th{\'e}o Moutakanni, Huy Vo, Marc
  Szafraniec, Vasil Khalidov, Pierre Fernandez, Daniel Haziza, Francisco Massa,
  Alaaeldin El-Nouby, Mahmoud Assran, Nicolas Ballas, Wojciech Galuba, Russell
  Howes, Po-Yao Huang, Shang-Wen Li, Ishan Misra, Michael Rabbat, Vasu Sharma,
  Gabriel Synnaeve, Hu~Xu, Herv{\'e} J{\'e}gou, Julien Mairal, Patrick Labatut,
  Armand Joulin, and Piotr Bojanowski.
\newblock {DINOv2}: Learning robust visual features without supervision.
\newblock \emph{arXiv preprint arXiv:2304.07193}, 2023.
\newblock URL \url{https://arxiv.org/abs/2304.07193}.

\bibitem[He et~al.(2025)He, Xu, Guo, Wetzstein, Dai, Li, and
  Yang]{he2025cameractrl}
Hao He, Yinghao Xu, Yuwei Guo, Gordon Wetzstein, Bo~Dai, Hongsheng Li, and
  Ceyuan Yang.
\newblock {CameraCtrl}: Enabling camera control for video diffusion models.
\newblock In \emph{International Conference on Learning Representations}, 2025.
\newblock URL
  \url{https://proceedings.iclr.cc/paper_files/paper/2025/hash/f98fd73d59d8494489ea970747b91fe4-Abstract-Conference.html}.

\bibitem[Wang et~al.(2024)Wang, Yuan, Wang, Li, Chen, Xia, Luo, and
  Shan]{wang2024motionctrl}
Zhouxia Wang, Ziyang Yuan, Xintao Wang, Yaowei Li, Tianshui Chen, Menghan Xia,
  Ping Luo, and Ying Shan.
\newblock {MotionCtrl}: A unified and flexible motion controller for video
  generation.
\newblock In \emph{ACM SIGGRAPH 2024 Conference Papers}, 2024.
\newblock \doi{10.1145/3641519.3657518}.

\bibitem[Shen et~al.(2026)Shen, Zhu, Li, Xie, Deka, Liu, Tang, Chen, and
  Zhang]{shen2026controllable}
Yifan Shen, Peiyuan Zhu, Zijian Li, Shaoan Xie, Namrata Deka, Zongfang Liu,
  Zeyu Tang, Guangyi Chen, and Kun Zhang.
\newblock Controllable video generation with provable disentanglement.
\newblock In \emph{International Conference on Learning Representations}, 2026.
\newblock URL \url{https://openreview.net/forum?id=OcLNKpcY4J}.

\bibitem[Lin et~al.(2023)Lin, Zhang, Chen, Pramanick, Gao, Wang, Yan, and
  Shou]{lin2023univtg}
Kevin~Qinghong Lin, Pengchuan Zhang, Joya Chen, Shraman Pramanick, Difei Gao,
  Alex~Jinpeng Wang, Rui Yan, and Mike~Zheng Shou.
\newblock {UniVTG}: Towards unified video-language temporal grounding.
\newblock In \emph{Proceedings of the IEEE/CVF International Conference on
  Computer Vision}, pages 2794--2804, 2023.

\bibitem[Wang et~al.(2025{\natexlab{d}})Wang, Wang, Xu, Du, Lin, Xiao, Yue, Ju,
  Zhang, Yang, Fang, He, Luo, Wang, Lin, Luan, and Jin]{wang2025timer1}
Ye~Wang, Ziheng Wang, Boshen Xu, Yang Du, Kejun Lin, Zihan Xiao, Zihao Yue,
  Jianzhong Ju, Liang Zhang, Dingyi Yang, Xiangnan Fang, Zewen He, Zhenbo Luo,
  Wenxuan Wang, Junqi Lin, Jian Luan, and Qin Jin.
\newblock {Time-R1}: Post-training large vision language model for temporal
  video grounding.
\newblock In \emph{Advances in Neural Information Processing Systems},
  volume~38, 2025{\natexlab{d}}.
\newblock \doi{10.52202/085713-2793}.

\end{thebibliography}

\newpage
\appendix
\onecolumn

{
    \centering
    \Large
    \textbf{MEV: A Multi-Event Video Dataset for Long-Take Generation} \\ \vspace{0.5em} {Supplementary Material}\\
    \large
    \par
}

\vspace{-0.1in}

\section*{}
\begin{center}
    \emph{\bf \em \large Outline}
\end{center}
{This supplementary material presents related work, the annotation taxonomy and field definitions of the MEV Dataset, annotation guidelines, data release and responsible-use details, experimental setup and results of our SFT model and other VLMs, additional dataset statistics and analysis, implementation details of MEV-Wan, the MEV-Eval protocol, additional qualitative results for multi-event video generation, and broader use and limitations.
\begin{itemize}
\item Section~\ref{sup:related} reviews related video-text datasets and multi-event video generation methods.
\item Section~\ref{sup:tax} provides annotation taxonomy and field definitions of MEV Dataset.
\item Section~\ref{sup:guide} details our annotation guidelines.
\item Section~\ref{sup:release} describes data sources, release format, licensing, and responsible-use considerations.
\item Section~\ref{sup:ex1} provides experimental setup and results of our SFT model and other VLMs.
\item Section~\ref{sup:stats} shows more statistics and analysis of MEV Dataset.
\item Section~\ref{sup:mevg} details the implementation of MEV-Wan for multi-event video generation.
\item Section~\ref{sup:mev_eval} describes the MEV-Eval protocol for multi-event video generation.
\item Section~\ref{sup:qual_gen} provides additional qualitative results and failure cases.
\item Section~\ref{sup:future_use} discusses broader use, limitations, and future directions.
\end{itemize}
}
\section{Related Work}
\label{sup:related}

\subsection{Video-Text Datasets}
Video–text datasets link natural videos with human language, offering paired temporal signals that have become central to recent progress in video understanding and generation. 
Pioneering efforts have curated a diverse array of video-text datasets tailored for specific tasks, encompassing action recognition~\citep{soomro2012ucf101,kay2017kinetics}, action temporal segmentation~\citep{tang2019coin,sener2022assembly101}, video-text retrieval~\citep{krishna2017dense}, and video question answering~\citep{mangalam2023egoschema}.
However, these resources vary substantially in visual quality, caption alignment, and temporal structure, making them difficult to directly leverage for learning high-quality video generation models. The emergence of large-scale video generation systems such as Sora~\citep{openai2024video} further suggests that the availability of high-quality video data and sufficient computational resources are crucial factors for advancing video generation.
Early web-scale datasets for text-to-video generation (\eg, HowTo100M~\citep{miech2019howto100m}, WebVid-10M~\citep{bain2021frozen}, HD-VILA-100M~\citep{xue2022advancing}, InternVid~\citep{wang2023internvid}, and Panda-70M~\citep{chen2024panda}) provide massive, largely weakly aligned video–text pairs that underpin large-scale representation pretraining and video–language modeling.
However, these datasets are typically dominated by short clips or loosely aligned web videos whose captions describe high-level situations rather than precise temporal evolution, offering limited supervision for long-horizon, multi-step generation.

To better support generation, recent datasets emphasize higher-quality captions and/or longer takes.
OpenVid-1M and OpenVidHD-0.4M~\citep{nan2024openvid} curate high-quality text-to-video pairs at scale for training diffusion-based generators, while VideoUFO~\citep{wang2025videoufo} focuses on user-centric prompts.
Complementary directions target richer long-form descriptions and longer durations with temporally dense and structured captions~\citep{xiong2024lvd,ju2024miradata,xue2025ultravideo,li2025openhumanvid}. Other works improve coverage of metric conditions~\citep{wang2025koala}, provide high-resolution videos with comprehensive captions~\citep{xue2025ultravideo}, or curate multi-source long-form captions with accessible data~\citep{yang2025vimix}.
While these datasets substantially improve caption fidelity and video quality, most still lack explicit segmentation into non-overlapping, temporally bounded events and do not provide paired global- and event-level descriptions tailored for multi-event generation.
In contrast, our dataset is designed to be subject-centric and multi-event, with precise temporal boundaries and procedural descriptions at both sequence and event granularity, enabling long-take generation and temporally targeted editing. Beyond global-level structured captions, we further provide event-level descriptions that explicitly consider factors such as subject appearance/disappearance and camera motion from multiple perspectives, better supporting fine-grained temporal control.

\subsection{Multi-Event Video Generation}
Recent progress in long-form video generation motivates multi-event video generation, where videos evolve through multiple actions and events, often forming coherent narratives.
The pioneering work Phenaki~\citep{villegas2022phenaki} first generates videos from time-varying event-level prompts.
MEVG~\citep{oh2024mevg} facilitates multi-event video generation by anchoring the current noise latent to the preceding clip’s terminal frame, ensuring visual coherence across semantic transitions.
More recently, mainstream approaches for multi-event video generation can be broadly grouped into two modeling paradigms. The first is temporal autoregressive generation, where videos are produced sequentially over time. Representative temporally autoregressive streaming diffusion methods use causal conditioning and cached past context to enable low-latency, extendable text-to-video synthesis~\citep{yin2024slow,henschel2025streamingt2v,huang2025self}. LongLive~\citep{yang2025longlive} further introduces a frame-level autoregressive framework for real-time interactive long video generation, enabling multi-event evolution via prompt switching.
The second paradigm is divide-and-conquer, motivated by the observation that strictly sequential generation can inevitably accumulate errors and degrade quality over long horizons. 
MinT~\citep{wu2025mind} binds each event to explicit timestamps and employs a rescaled rotary positional encoding to enable fine-grained control over event timing within a single diffusion-based model.
SynCoS~\citep{kim2025tuning} instead synchronizes DDIM~\citep{song2020denoising}-based local fusion and CSD~\citep{kim2023collaborative}-based global optimization via grounded timesteps and fixed baseline noise to ensure smooth transitions.
Related but distinct, multi-shot video generation targets a sequence of discontinuous yet semantically connected shots~\citep{kara2025shotadapter,wang2025echoshot,an2025onestory}. The multi-shot property is orthogonal to the number of events, since even a single-event narrative can be rendered as multiple shots by varying cinematic viewpoints.

\section{Annotation Taxonomy and Field Definitions}
\label{sup:tax}

MEV adopts a hierarchical annotation taxonomy that combines global video-level descriptions with fine-grained event-level annotations. At the global level, each video is described by 8 structured fields that cover persistent scene properties, namely background, theme, style, shot type, camera movement, lighting, atmosphere, and subject profile, as summarized in Table~\ref{tab:global_structured_caption}. In parallel, we provide 3 multi-grained free-form summaries, i.e., short, middle, and long captions, so that each video can be represented at different levels of semantic detail, as shown in Table~\ref{tab:summary_captions}.

At the event level, MEV annotates semantic and camera-related attributes to describe local changes, participant composition, and activity or scene context. The semantic fields include appearance, disappearance, environmental change, and lighting change; multi-person, animal, and child presence; and cooking, sports, and travel (Table~\ref{tab:semantic_attributes}). These non-exclusive fields support subset retrieval and targeted training, with detailed actions described by the event captions. Camera-related annotations further describe translation, rotation, zoom, and focus shift, which together provide a more complete representation of how the scene evolves over time. This structured taxonomy supports both annotation consistency and downstream modeling for multi-event video understanding and generation.

For the camera taxonomy, we follow the basic decomposition of camera motion in CameraBench~\citep{lin2025towards}, but only retain the most common and visually reliable motion primitives in real-world videos. In our dataset, static videos still account for the majority, so we prioritize a simplified label space centered on frequent motions such as dolly, truck, pedestal, pan, tilt, roll, and zoom, as detailed in Tables~\ref{tab:camera_translation}--\ref{tab:camera_zooming}. Focus shifts are recorded separately as natural-language descriptions (Table~\ref{tab:focus_shift}). We do not further subdivide more complex cases such as object-centric tracking variants, motion speed, or cinematic effects. Instead, these finer-grained details are described in the camera motion description field. This design is more practical for VLM-based annotation, since many such cases require jointly reasoning about camera trajectory, subject motion, and relative viewpoint over time. This simplified camera taxonomy has three practical advantages. First, it still covers a large portion of real-world camera motion patterns in MEV. Second, the retained labels are more distinguishable for the VLM annotator, leading to more stable automatic annotation. Third, the labels are atomic and compositional, so multiple camera labels can be assigned simultaneously when needed, such as {\tt pan-right} together with {\tt tilt-up}.

\begin{table*}[h!]
\centering
\caption{\small \textbf{Global structured caption} fields used in MEV.}
\scalebox{0.8}{
\begin{NiceTabular}{M{0.28\linewidth} M{0.62\linewidth}}
\CodeBefore
    \Body
\toprule[1.2pt]
\textbf{Structure} & \textbf{Definition} \\
\midrule
\textbf{Background} & Describes the physical setting, surrounding environment, and salient background elements that remain relevant across the whole video. \\
\midrule
\textbf{Theme} & Summarizes the central idea, activity, or story premise of the clip at a high semantic level. \\
\midrule
\textbf{Style} & Specifies the aesthetic treatment, cinematic style, or realism level of the video. \\
\midrule
\textbf{Shot Type} & Defines the camera framing, shot scale, and viewpoint, such as close-up, wide shot, over-the-shoulder, or low-angle shot. \\
\midrule
\textbf{Camera Movement} & Describes the dominant camera motion pattern throughout the clip, including whether the camera is static or moving and the major motion direction. \\
\midrule
\textbf{Lighting} & Characterizes the lighting condition, including source, intensity, color tone, and time-of-day cues. \\
\midrule
\textbf{Atmosphere} & Captures the overall emotional tone or ambience conveyed by the scene. \\
\midrule
\textbf{Subject Profile} & Provides a persistent description of the main subjects, including their appearance, role, clothing, and other stable attributes. \\
\bottomrule[1.2pt]
\end{NiceTabular}
}
\label{tab:global_structured_caption}
\end{table*}

\begin{table*}[h!]
\centering
\caption{\small \textbf{Multi-grained global summary captions} used in MEV.}
\scalebox{0.8}{
\begin{NiceTabular}{M{0.26\linewidth} M{0.56\linewidth}}
\CodeBefore
    \Body
\toprule[1.2pt]
\textbf{Caption Level} & \textbf{Definition} \\
\midrule
\textbf{Short Caption} & Provides a compact one-sentence summary of the overall video, emphasizing the main action and subjects, typically in around 20 words. \\
\midrule
\textbf{Middle Caption} & Gives a moderately detailed summary of the full clip, covering key subjects, actions, scene context, and overall visual impression, typically in around 80 words. \\
\midrule
\textbf{Long Caption} & Provides a detailed holistic description of the video, including subject attributes, environment, motion, camera behavior, and atmosphere, typically in around 200 words. \\
\bottomrule[1.2pt]
\end{NiceTabular}
}
\label{tab:summary_captions}
\end{table*}

\begin{table*}[h!]
\centering
\caption{\small \textbf{Local event semantic attributes} used in MEV.}
\scalebox{0.83}{
\begin{NiceTabular}{M{0.23\linewidth} M{0.25\linewidth} M{0.4\linewidth}}
\CodeBefore
    \Body
\toprule[1.2pt]
\textbf{Attribute} & \textbf{Field} & \textbf{Definition} \\
\midrule
\textbf{Appearance} & {\tt has\_appearance} & Indicates that a subject or salient object newly enters the frame or becomes newly visible during the event. \\
\midrule
\textbf{Disappearance} & \makecell[l]{\tt has\_\\\tt disappearance} & Indicates that a subject or salient object exits the frame or becomes no longer visible during the event. \\
\midrule
\textbf{Multi-person} & {\tt is\_multi\_person} & Indicates that more than one person is actively present or involved in the event. \\
\midrule
\textbf{Environmental change} & \makecell[l]{\tt has\_\\\tt environmental\_\\\tt change} & Indicates a noticeable change in the scene state or background context beyond ordinary subject motion. Newly revealed content caused only by camera motion or frame expansion does not count as environmental change; instead, the environment itself must change, such as leaves falling in the background or a car passing through the scene. \\
\midrule
\textbf{Lighting change} & \makecell[l]{\tt has\_lighting\_\\change} & Indicates a salient change in illumination, exposure, or lighting condition within the event. \\
\midrule
\textbf{Animal} & {\tt has\_animal} & Indicates that an animal is present as a meaningful semantic element of the event. \\
\midrule
\textbf{Cooking} & {\tt has\_cooking} & Indicates that the event centers on cooking, food preparation, or a closely related kitchen activity. \\
\midrule
\textbf{Sports} & {\tt has\_sports} & Indicates that the event contains sports- or fitness-related activity. \\
\midrule
\textbf{Travel} & {\tt has\_travel} & Indicates that the event involves travel- or tourism-related activity or context. \\
\midrule
\textbf{Child presence} & {\tt has\_child} & Indicates that a child is present in the event. \\
\bottomrule[1.2pt]
\end{NiceTabular}
}
\label{tab:semantic_attributes}
\end{table*}

\begin{table*}[h!]
\centering
\caption{\small \textbf{Camera translation} definitions and guidelines.}
\scalebox{0.8}{
\begin{NiceTabular}{l M{0.3\linewidth} M{0.6\linewidth}}
\CodeBefore
    \Body
\toprule[1.2pt]
\textbf{Translation} & \textbf{Options} & \textbf{Definition} \\
\midrule
\multirow{2}{*}{\textbf{Dolly}}
& {\tt dolly-in/\newline dolly-out} & The camera moves forward or backward relative to the ground plane and the initial frame.\\\cmidrule{2-3}
& {\tt no-dolly} & The camera does not move forward/backward during the shot.\\ \midrule
\multirow{3}{*}{\textbf{Pedestal}}
& {\tt pedestal-up/\newline pedestal-down}  & Select this when the camera moves upward or downward clearly and consistently relative to the ground or the orientation of the initial frame. \\ \cmidrule{2-3}
& {\tt no-pedestal} & Select this label when the camera does not move upward/downward during the shot. \\ \midrule
\multirow{2}{*}{\textbf{Truck}}
& {\tt truck-left/\newline truck-right}  & The camera physically moves to the left or right, changing its position relative to the initial frame.  \\\cmidrule{2-3}
& {\tt no-truck} & The camera does not move to the left or right during the shot.\\
\bottomrule[1.2pt]
\end{NiceTabular}
}
\label{tab:camera_translation}
\end{table*}

\begin{table*}[h!]
\centering
\caption{\small \textbf{Camera rotation} definitions and guidelines.}
\scalebox{0.8}{
\begin{NiceTabular}{l M{0.3\linewidth} M{0.6\linewidth}}
\CodeBefore
    \Body
\toprule[1.2pt]
\textbf{Rotation} & \textbf{Options} & \textbf{Definition} \\ \midrule
\multirow{2}{*}{\textbf{Pan}}
& {\tt pan-left/\newline pan-right}  & The camera rotates its angle by pivoting left or right with respect to the initial frame.  \\\cmidrule{2-3}
& {\tt no-pan} & The camera does not pan left or right.\\
\midrule
\multirow{3}{*}{\textbf{Tilt}}
& {\tt tilt-up/\newline tilt-down}  & The camera rotates its angle up or down vertically with respect to the initial frame. \\ \cmidrule{2-3}
& {\tt no-tilt} & The camera does not tilt up or down. \\ \midrule
\multirow{4}{*}{\textbf{Roll}}
& {\tt roll-CW/\newline roll-CCW}  & The camera performs a clear and consistent clockwise (CW) or counterclockwise (CCW) roll by rotating around its own optical center. \\\cmidrule{2-3}
& {\tt no-roll} & The camera does not roll clockwise/counterclockwise. \\
\bottomrule[1.2pt]
\end{NiceTabular}
}
\label{tab:camera_rotation}
\end{table*}

\begin{table*}[h!]
\centering
\caption{\small \textbf{Camera intrinsic change} definitions and guidelines.}
\scalebox{0.8}{
\begin{NiceTabular}{l M{0.3\linewidth} M{0.6\linewidth}}
\CodeBefore
    \Body
\toprule[1.2pt]
\textbf{Zooming} & \textbf{Options} & \textbf{Definition} \\
\midrule
\multirow{2}{*}{\textbf{Zoom}}
& {\tt zoom-in/\newline zoom-out} & The camera adjusts its focal length to zoom in or out, changing the frame size. Note: This differs from physical camera movement. \\\cmidrule{2-3}
& {\tt no-zoom} & The camera does not adjust its focal length during the video.\\
\bottomrule[1.2pt]
\end{NiceTabular}
}
\label{tab:camera_zooming}
\end{table*}

\begin{table*}[h!]
\centering
\caption{\small \textbf{Focus shift} field and annotation guidelines.}
\scalebox{0.8}{
\begin{NiceTabular}{l M{0.3\linewidth} M{0.6\linewidth}}
\CodeBefore
    \Body
\toprule[1.2pt]
\textbf{Field} & \textbf{Value / format} & \textbf{Definition} \\
\midrule
\multirow{2}{*}{\texttt{focus\_shift}}
& \texttt{from ... to ...} & Describe a visible transfer of optical focus, reflected in changes in local sharpness across subjects or regions, \eg, ``from the woman in the background to the man in the foreground''. \\\cmidrule{2-3}
& Empty value & No focus shift occurs within the event. A change in the subject shown due only to camera movement or reframing does not count as a focus shift. \\
\bottomrule[1.2pt]
\end{NiceTabular}
}
\label{tab:focus_shift}
\end{table*}

\section{Annotation Guidelines}
\label{sup:guide}

\begin{figure*}[t]
    \centering
    \includegraphics[width=1.0\linewidth]{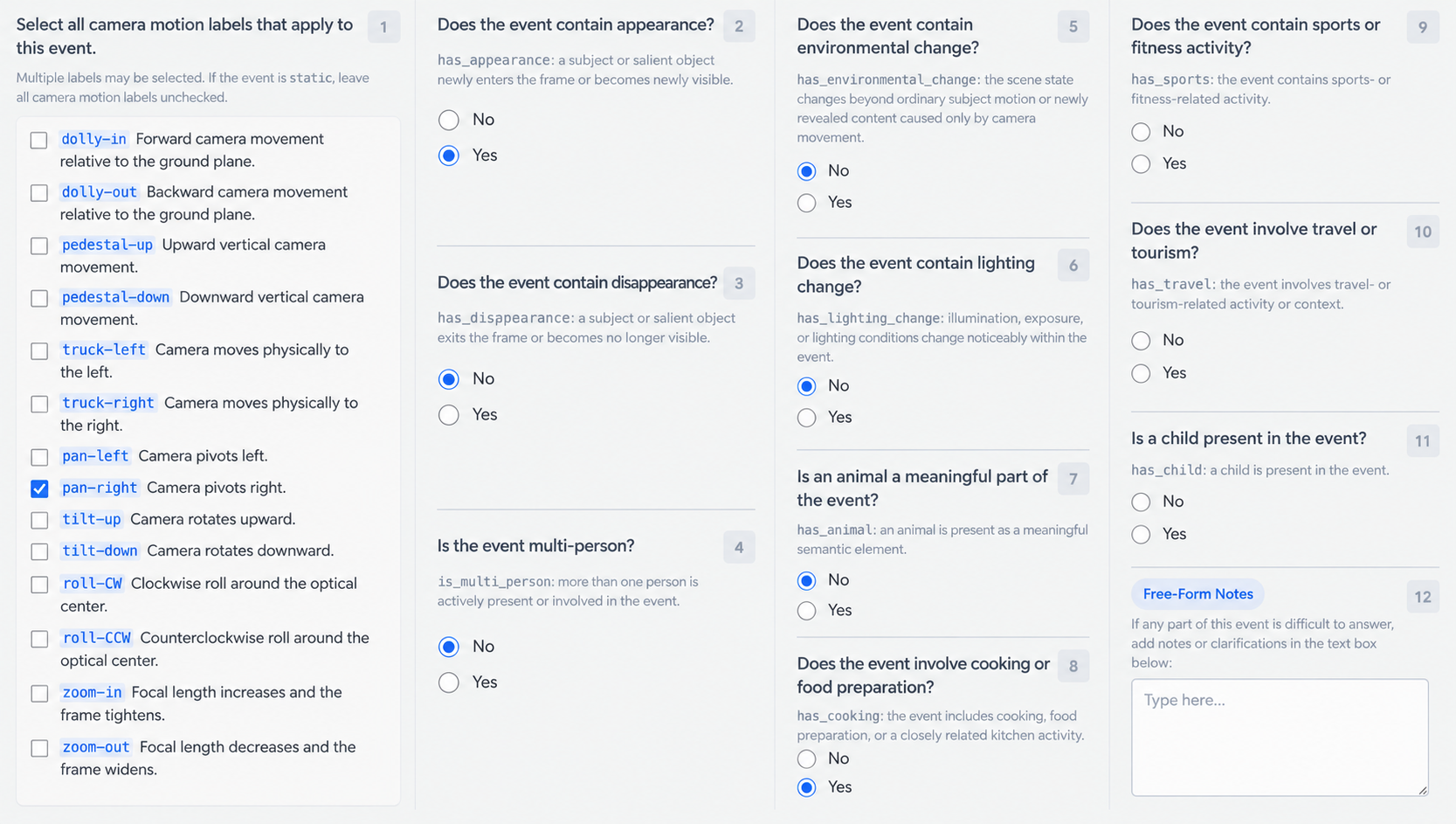}
    \caption{Example questions in our annotation framework.}
    \label{fig:survey}
\end{figure*}

\noindent \textbf{Annotation interface.}
Our annotation workflow combines a question-based labeling interface with a lightweight temporal segmentation tool. Figure~\ref{fig:survey} shows example questions in our annotation framework. Annotators watch the full video, answer event-level questions, and revise previous selections whenever later observations reveal a better interpretation. In practice, this interface is used after temporal segmentation to annotate whether each event contains camera motion and semantic attributes, while the free-form camera motion description field is used to record finer details that are not explicitly enumerated in our discrete labels. This design helps reduce contradictory labels and keeps the annotation process consistent with the taxonomy in Section~\ref{sup:tax}.

\begin{figure*}[t]
    \centering
    \includegraphics[width=1.0\linewidth]{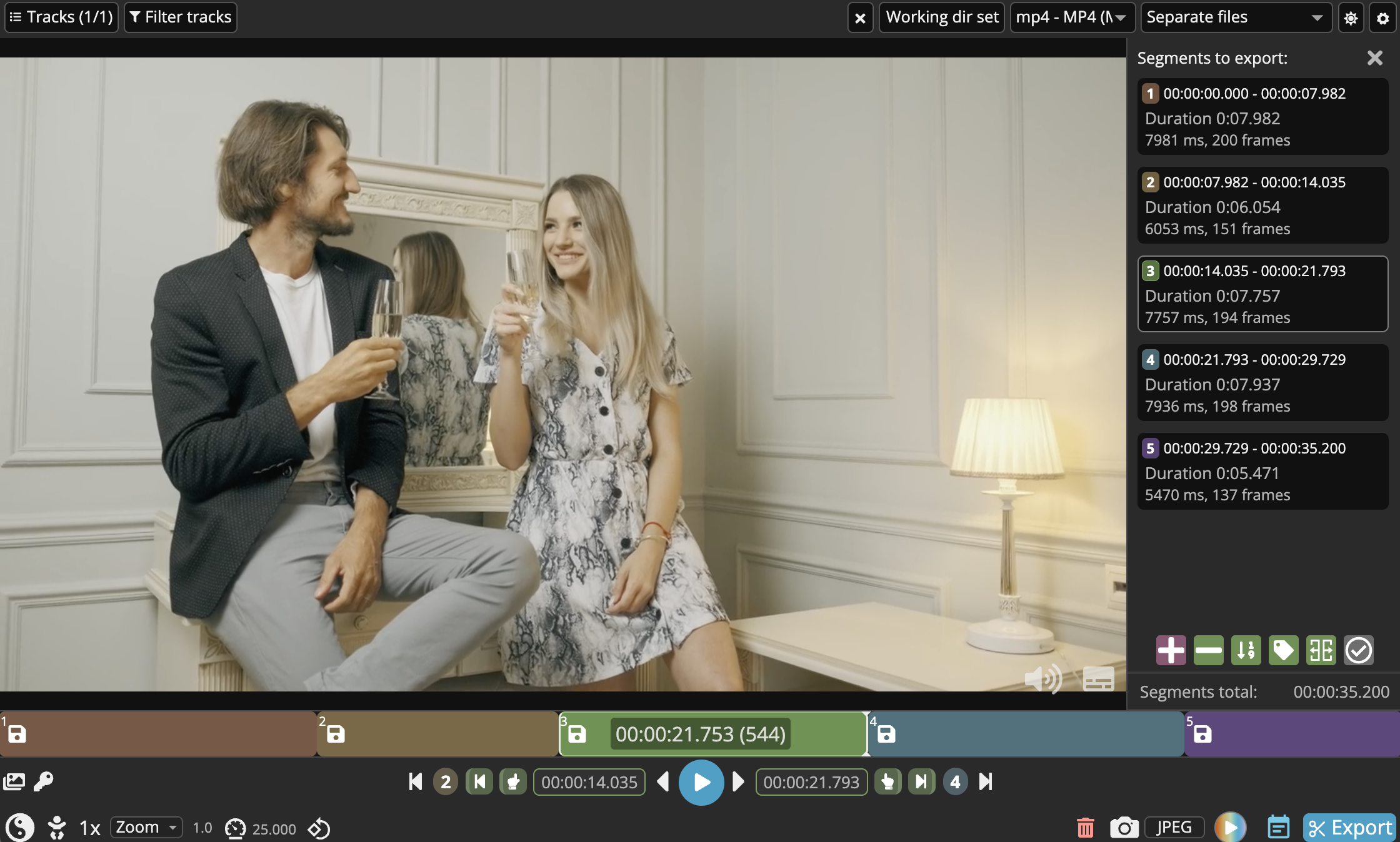}
    \caption{Annotating multi-event temporal boundaries based on LosslessCut.}
    \label{fig:losslesscut}
\end{figure*}

\noindent \textbf{Temporal boundary annotation based on LosslessCut.}
For multi-event videos, we first annotate temporal boundaries with LosslessCut, as illustrated in Figure~\ref{fig:losslesscut}. The workflow is: (1) import the original video into LosslessCut; (2) play the video and observe activity changes over time; (3) based on the event segmentation rules in Section~\ref{mev_step_3}, place a cut marker whenever a new event begins; and (4) after all boundaries are marked, use the export function to automatically obtain all sub-videos together with their begin and end timestamps. After segmentation, annotators further label each event with camera motion and semantic attributes when applicable. This decomposition is practical because it separates temporal grounding from semantic labeling, making the overall process easier to review, correct, and scale.

\noindent \textbf{Human consensus and dataset triage.}
We employ three annotators to inspect videos from the curated pool and decide whether each sample should be kept as a multi-event video or filtered out as single-event or difficult. In practice, different annotators naturally have different sensitivity to small action changes, so disagreements often arise not because the video is ambiguous everywhere, but because one annotator prefers finer decomposition than another. To make the decision reproducible, we only keep videos whose event structure remains stable across annotators. For temporal boundaries, we treat two timestamps $t_1$ and $t_2$ as matching when $|t_1 - t_2| \le \delta$, where we set $\delta=0.5$ seconds. Empirically, this threshold is stable: it is large enough to absorb minor reaction-time differences in manual clicking, yet small enough to preserve event-level precision in long-take videos. If at least two of the three annotators agree under this criterion, we adopt one of the agreed timestamps; otherwise, we either merge neighboring segments or filter the sample if no stable decomposition can be reached.

This consensus rule is especially important because MEV favors action-driven atomicity rather than over-segmentation. When one annotator proposes more events but the additional cuts are not reproducible across the other annotators, we do not keep the finer split. Instead, we choose the more stable segmentation that can be consistently named and grounded by multiple people. In addition, a video is still treated as single-event when one coherent activity is accompanied by one consistent camera motion throughout the clip. By contrast, if the camera behavior itself changes in a semantically meaningful way and forms a new temporally grounded stage---for example, a static landscape shot followed by a {\tt pan-right} reveal---we treat this transition as a new event.

Most candidate videos are eventually filtered out at this stage. Many clips are too short, contain only one coherent event, or lack consistently nameable stages and reproducible boundaries. Continuous sports or performance motions are assigned to the difficult category only when they fail these event-segmentation criteria. As a result, only a relatively small portion of the large curated pool is retained for detailed event-level annotation. Following the annotation split in Section~\ref{mev_step_3}, the manual inspection stage covers 52.9\% of the curated pool, and only samples retained as reliable multi-event videos receive full event-level timestamps, captions, camera descriptions, and semantic attributes.

\noindent \textbf{Agreement statistics.}
Table~\ref{tab:human_triage_agreement} reports agreement at two levels, both aligned with the $\delta=0.5$s timestamp-matching rule in Section~\ref{mev_step_3}. Triage agreement measures the outcome of the three-annotator video-level decision before detailed event annotation. Boundary reconciliation is measured only on retained multi-event candidates and captures whether timestamp proposals require consensus alignment before finalizing the segmentation.

\begin{table}[h]
\centering
\caption{\small \textbf{Human triage and boundary agreement statistics.}}
\begin{tabular}{>{\raggedright\arraybackslash}p{0.38\linewidth} c >{\raggedright\arraybackslash}p{0.40\linewidth}}
\toprule[1.2pt]
\textbf{Statistic} & \textbf{Ratio} & \textbf{Denominator and meaning} \\
\midrule
Unanimous single-event decision & 51.2\% & Manually inspected candidates; all annotators judge no stable multi-event structure \\
Difficult / ambiguous decision & 27.7\% & Manually inspected candidates; event decomposition is unreliable or hard to name consistently \\
Retained multi-event decision & 21.1\% & Manually inspected candidates; clip passes reproducible event-structure triage \\
Boundary reconciliation & 97.7\% & Retained multi-event candidates; timestamps require consensus alignment before finalization \\
\bottomrule[1.2pt]
\end{tabular}
\label{tab:human_triage_agreement}
\end{table}

\section{Data Sources, Release, Licensing, and Responsible Use}
\label{sup:release}

\noindent \textbf{Source selection and scope.}
MEV is constructed from public video sources, primarily MixKit and a complementary set of YouTube-hosted videos. MixKit provides visually clean stock-style videos with stable imaging quality, while YouTube complements it with long-tail activity patterns that are less frequent in stock footage. For YouTube candidates, we use predefined activity-oriented categories rather than unrestricted crawling. These categories focus on visually clear long-take activities such as daily activities, cooking, fitness or sports, tool use or DIY, pet or animal interactions, and outdoor activities. We then apply the filtering and event-triage pipeline described in Sections~\ref{mev_step_12} and~\ref{mev_step_3} to retain only clips that satisfy MEV's long-take multi-event requirements.

\noindent \textbf{Dataset availability.}
The MEV dataset is publicly available at \url{https://huggingface.co/datasets/MultiEventVideo/MEV}, including the official release package and accompanying metadata.

\noindent \textbf{Sensitive-content filtering and bias.}
MEV is intended for research on long-take multi-event generation. During collection and filtering, we exclude videos centered on sensitive or private-context content, including real-world violence, accidents, injuries, medical or hospital scenes, political or religious events, adult or explicit sexual content, surveillance or private security footage, hate or extremist content, news, and other highly sensitive public events. We also avoid annotating identity or sensitive personal attributes. These choices make MEV biased toward visually clear, action-rich, non-sensitive clips, which is appropriate for the target generation setting but should be considered when interpreting dataset coverage.

\noindent \textbf{Licensing.}
MEV is released as an open research dataset. Author-created MEV annotations, labels, timestamps, prompts, and evaluation metadata are released under the Creative Commons Attribution 4.0 International License (CC BY 4.0), unless a file explicitly states otherwise. Video files included in the official public release are provided together with MEV annotations as a directly usable research data package for evaluation and model training, subject to the applicable source license or platform terms. MEV does not impose additional restrictions beyond those source terms and responsible-use requirements, and users do not need to re-crawl source websites for videos included in the official package. Only a small subset of optional 4K source versions may be premium, paid, or otherwise restricted; these optional variants are not part of the public package, and MEV provides annotations, timestamps, prompts, and other metadata where permitted. Full release details are provided in the repository license notice.

\noindent \textbf{Takedown policy.}
Users should comply with the applicable source terms. If a rights holder requests removal of a video file, video reference, annotation, or related metadata, or if legal, ethical, privacy, or security concerns arise, we may remove or modify the affected materials. Affected videos are removed from subsequent public packages when appropriate.

\section{Experiments on Annotation Model}
\label{sup:ex1}
\noindent \textbf{VLM Details.}
To improve automatic event segmentation, we fine-tune an event-aware Qwen3-VL-32B model on SFT data constructed from the manually labeled videos described in Section~\ref{mev_step_3}. In particular, the temporally grounded multi-event portion contains about 5K videos with human-annotated event boundaries, which serve as the core supervision for boundary prediction. We serialize the supervision into a small set of instruction-following video-text templates that mirror our annotation workflow. As summarized in Table~\ref{tab:sft_formats}, the SFT data covers three practical forms: a status-only template for identifying videos that should be filtered out as {\tt single-event} or {\tt difficult}, a timestamp template for retained {\tt multi-event} videos, and an explanation template for borderline cases where a short textual reason is useful for later human review. This design keeps the output easy to parse while remaining close to the way human annotators reason about event segmentation. For Qwen3-VL-32B~\citep{Qwen3-VL}, we adopt an LLM-only supervised fine-tuning setup: the language model is trainable, while the vision tower and the multimodal projector remain frozen. We sample videos at 8 FPS and train with bf16, cosine learning-rate decay, a warmup ratio of 0.05, FlashAttention-2, and DeepSpeed ZeRO-3. Table~\ref{tab:qwen3_sft_hyperparams} summarizes the configuration used for our 32B model.

\begin{table}[h]
\centering
\caption{\small \textbf{SFT data formats} used for event segmentation.}
\begin{tabular}{p{0.23\linewidth} p{0.28\linewidth} p{0.36\linewidth}}
\toprule[1.2pt]
\textbf{Format} & \textbf{Input} & \textbf{Target Output} \\
\midrule
\textbf{Status-only} & Video + segmentation instruction & {\tt single-event} or {\tt difficult} \\
\midrule
\textbf{Timestamped multi-event} & Video + segmentation instruction & {\tt multi-event} with an ordered list of event begin/end timestamps \\
\midrule
\textbf{Explanation-assisted} & Video + segmentation instruction & Filtering decision or merge decision with a short textual explanation for later review \\
\bottomrule[1.2pt]
\end{tabular}
\label{tab:sft_formats}
\end{table}

\begin{table}[h]
\centering
\caption{\small \textbf{SFT hyperparameters} for Qwen3-VL-32B.}
\begin{tabular}{c|c}
\toprule[1.2pt]
\textbf{Hyperparameter} & \textbf{Value} \\
\hline
tuning\_mode & LLM only \\
per\_device\_train\_batch\_size & 1 \\
gradient\_accumulation\_steps & 2 \\
learning\_rate & 2.0e-6 \\
num\_train\_epochs & 6.0 \\
lr\_scheduler\_type & cosine \\
warmup\_ratio & 0.05 \\
tune\_mm\_llm & true \\
tune\_mm\_mlp & false \\
tune\_mm\_vision & false \\
video\_fps & 8.0 \\
video\_max\_pixels & 16384 \\
image\_max\_pixels & 262144 \\
deepspeed & zero3\_config.json \\
template & qwen3\_vl \\
bf16 & true \\
flash\_attn & fa2 \\
\bottomrule[1.2pt]
\end{tabular}
\label{tab:qwen3_sft_hyperparams}
\end{table}

\noindent \textbf{Baselines.}
We compare our SFT model with three strong VLM baselines: Qwen3-VL-8B~\citep{Qwen3-VL}, Qwen3-VL-32B~\citep{Qwen3-VL}, and Gemini 2.5 Pro~\citep{Comanicietal2025}. Qwen3-VL-8B and Qwen3-VL-32B represent open-source models with different parameter scales, while Gemini 2.5 Pro serves as a strong proprietary baseline.

\noindent \textbf{Evaluation Metrics.}
Our task is a joint event counting and boundary prediction problem: given a video, the model predicts both the number of events and the temporal boundaries between adjacent events. We therefore report three complementary groups of metrics. First, for \emph{boundary detection}, we compare the predicted internal boundary set $B_P=\{\hat e_1,\dots,\hat e_{N-1}\}$ with the ground-truth set $B_G=\{e_1,\dots,e_{M-1}\}$ under a tolerance window $\delta$. A predicted boundary is counted as correct if $|\hat e-e|\le\delta$ and neither side has been matched before; unmatched predictions are false positives and unmatched ground-truth boundaries are false negatives. We report boundary precision, recall, and F1, with $\delta=0.5$s and $1.0$s, and use the 0.5s setting as the main number to stay consistent with the annotation protocol in Section~\ref{mev_step_3}.

Second, for \emph{segment-level overlap}, we compare the predicted segments $P=\{p_j\}_{j=1}^{N}$ and the ground-truth segments $G=\{g_i=[s_i,e_i]\}_{i=1}^{M}$ using temporal IoU. For a predicted interval $[s_p,e_p]$ and a ground-truth interval $[s_g,e_g]$, we define
\[
\mathrm{tIoU}=\frac{|[s_p,e_p]\cap[s_g,e_g]|}{|[s_p,e_p]\cup[s_g,e_g]|}.
\]
We compute one-to-one matches with maximum total overlap and report Segment F1 at tIoU$=0.7$, together with a length-weighted matched tIoU. These metrics evaluate whether the predicted temporal partition resembles the full ground-truth event structure, rather than only checking isolated cut points.

Third, because the model also predicts the event count, we separately report \emph{counting error}. Specifically, we use the absolute count error $|N-M|$ and the relative count error $|N-M|/M$. These metrics quantify event-count discrepancies that may not be apparent from boundary metrics alone.

\noindent \textbf{Experimental Results.}
We evaluate event counting and boundary prediction on videos from the remaining candidate pool, using human-verified annotations as references. Tables~\ref{tab:count_error_results} and~\ref{tab:boundary_segment_results} show that the SFT model consistently improves event segmentation quality over all baselines. It achieves the lowest counting error and the best boundary and segment-level performance, indicating that supervised adaptation not only reduces over- or under-segmentation, but also produces a temporal partition that is closer to the full ground-truth event structure.
Among the non-SFT baselines, Gemini 2.5 Pro and Qwen3-VL-32B are broadly comparable. Gemini shows slightly stronger recall-oriented behavior, leading to better boundary recall and slightly higher F1 under the looser matching setting, whereas Qwen3-VL-32B is slightly stronger in precision-oriented boundary detection and segment overlap quality. Qwen3-VL-8B is clearly weaker than the larger models across both counting and temporal localization metrics, confirming that model scale remains important for reliable long-video event segmentation.

\begin{table}[h]
\centering
\caption{\small \textbf{Count error results} for event segmentation. Lower is better.}
\begin{tabular}{lcc}
\toprule[1.2pt]
\textbf{Model} & \textbf{Mean $|N-M|$} & \textbf{Rel. Error} \\
\midrule
Qwen3-VL-8B & 1.24 & 0.41 \\
Qwen3-VL-32B & 0.86 & 0.29 \\
Gemini 2.5 Pro & 0.82 & 0.28 \\
Qwen3-VL-32B (SFT) & \textbf{0.59} & \textbf{0.20} \\
\bottomrule[1.2pt]
\end{tabular}
\label{tab:count_error_results}
\end{table}

\begin{table*}[h]
\centering
\caption{\small \textbf{Boundary and segment-level evaluation} for event segmentation. Higher is better.}
\begin{adjustbox}{max width=\linewidth}
\begin{tabular}{lcccccc}
\toprule[1.2pt]
\textbf{Model} & \textbf{B-P@0.5s} & \textbf{B-R@0.5s} & \textbf{B-F1@0.5s} & \textbf{B-F1@1s} & \textbf{Seg-F1@0.7} & \textbf{LW-mtIoU} \\
\midrule
Qwen3-VL-8B & 40.2 & 37.4 & 38.7 & 50.1 & 35.4 & 0.46 \\
Qwen3-VL-32B & 50.4 & 45.6 & 47.9 & 60.8 & 46.5 & 0.58 \\
Gemini 2.5 Pro & 48.9 & 48.3 & 48.6 & 61.7 & 45.9 & 0.57 \\
Qwen3-VL-32B (SFT) & \textbf{56.8} & \textbf{54.9} & \textbf{55.8} & \textbf{68.1} & \textbf{53.4} & \textbf{0.64} \\
\bottomrule[1.2pt]
\end{tabular}
\end{adjustbox}
\label{tab:boundary_segment_results}
\end{table*}

\begin{figure*}[t]
    \centering
    \includegraphics[width=1.0\linewidth]{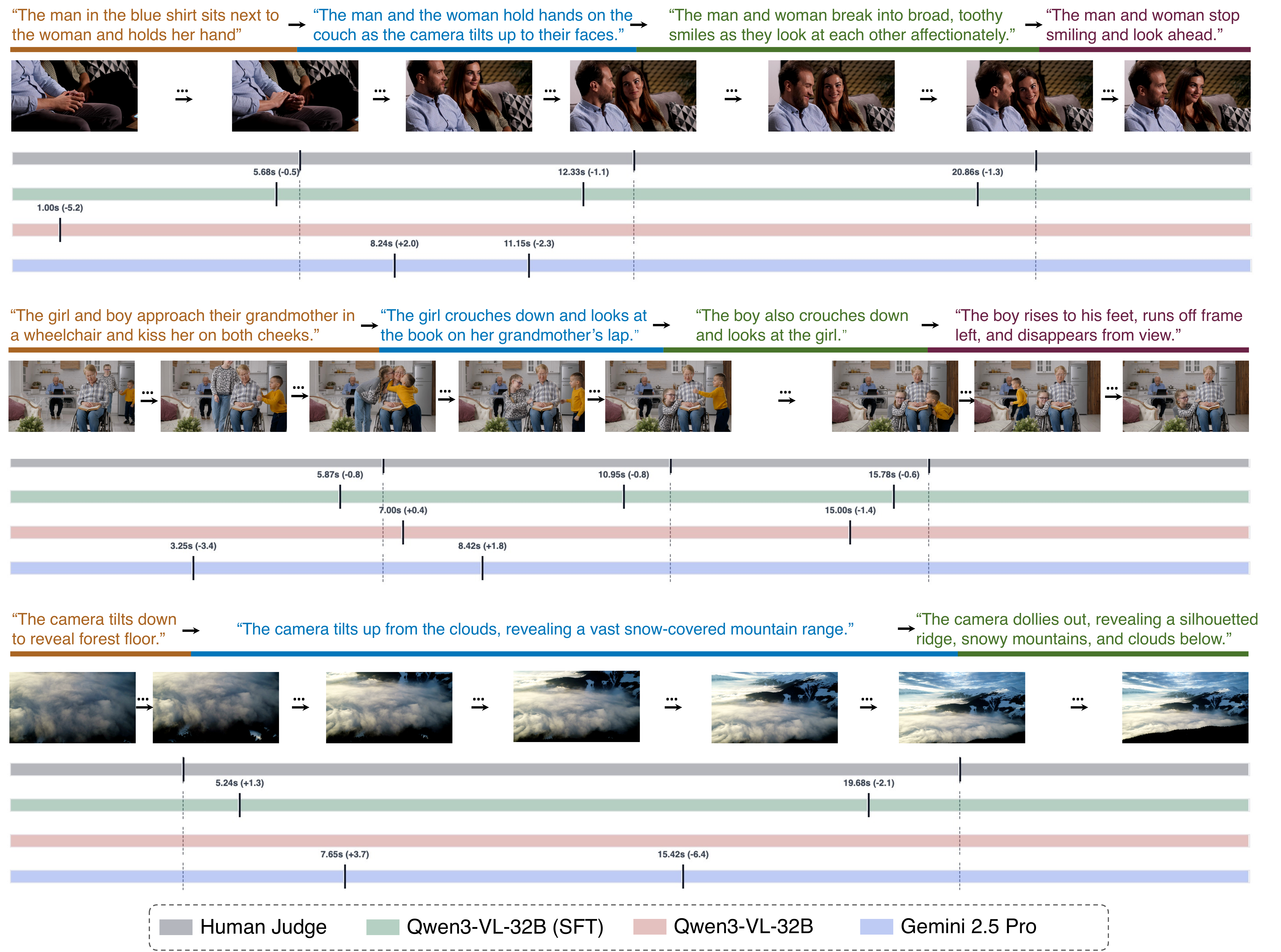}
    \caption{\small \textbf{Qualitative comparison of temporal boundary annotation.} Human annotations provide more atomic and temporally focused event windows than non-SFT VLM outputs. Qwen3-VL-32B (SFT) improves event-count recovery, but human verification remains important for boundary timing, camera-transition boundaries, and subject disappearance from view.}
    \label{comparison_temporal_boundary}
\end{figure*}

\noindent \textbf{Qualitative comparison of temporal boundary annotation.}
Figure~\ref{comparison_temporal_boundary} provides a qualitative comparison of human annotation, Qwen3-VL-32B (SFT), Qwen3-VL-32B, and Gemini 2.5 Pro on representative temporal segmentation cases. The SFT model is effective in recovering the coarse event structure and event count in many examples, matching the human annotation more closely than the non-SFT baselines. By contrast, native Qwen3-VL often fails to identify camera-behavior changes in long takes as valid event boundaries, and tends to aggregate different actions and camera-motion cues into a single broad event. However, the SFT model's timestamp precision is still imperfect. In the landscape example, the transition from a cloud-focused aerial view to the later camera movement that reveals the mountain range and foreground ridge is visually salient, but the SFT boundaries still deviate from the human boundaries. This indicates that supervised VLM segmentation is useful for scaling annotation, while human verification remains necessary for precise temporal grounding.

The examples also illustrate why human annotation can differ from VLM outputs in event count and granularity. MEV follows the event definition in Section~\ref{mev_step_3}: boundaries are not merely low-level motion changes, but correspond to stable changes in action state, scene state, interaction target, subject visibility, or camera behavior that can be named and grounded reproducibly. In the first example, the human boundaries separate a hand-centered static close-up, a camera tilt that moves the visual focus upward, a face-centered affectionate gaze, and a final phase in which both stop smiling and look ahead. The change between static and moving camera behavior therefore serves as a meaningful event cue rather than incidental visual motion, reducing ambiguity between physical interaction, camera transition, and social exchange. In the second example, human annotators isolate the window in which the boy runs around the wheelchair and disappears from the field of view, forming a complete event centered on subject disappearance. The non-SFT VLM captions describe the sustained family interaction but miss this visibility change, whereas the human-defined window makes the cue explicit and encourages more focused event captions. These atomic, modular event windows reduce boundary ambiguity and provide cleaner conditioning units for downstream generation, where explicit event structure helps the model follow localized actions instead of a coarse entangled description.

\section{Additional Analysis of MEV Dataset}
\label{sup:stats}

\begin{figure*}[t]
    \centering
    \includegraphics[width=1.0\linewidth]{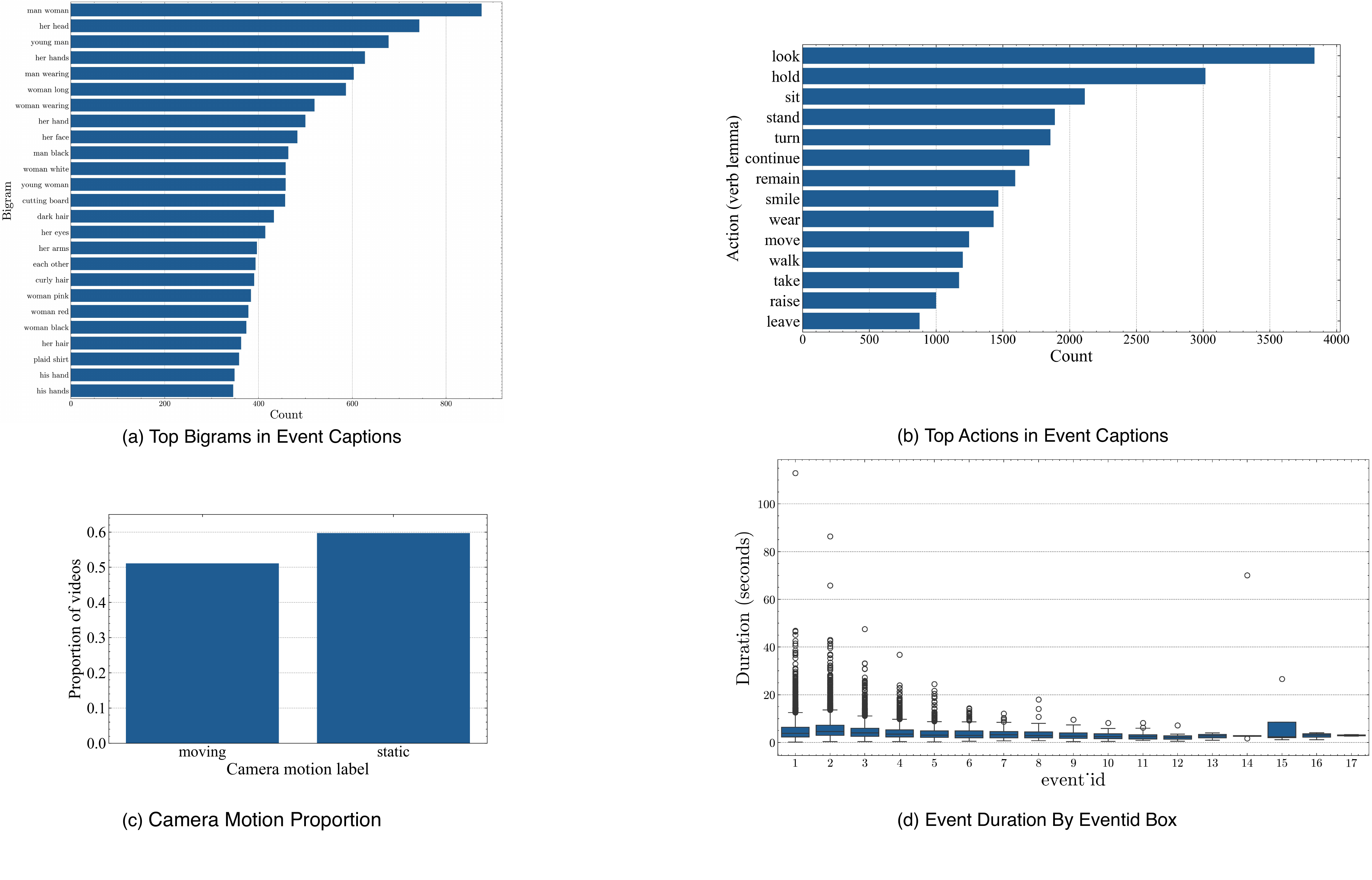}
    \caption{Additional statistics and analysis of MEV Dataset.}
    \label{fig:sup_stats}
\end{figure*}

\noindent \textbf{Additional dataset statistics.}
Figure~\ref{fig:sup_stats} provides four complementary views of the linguistic and temporal structure of MEV. Figure~\ref{fig:sup_stats}(a) shows the top bigrams in event captions. The most frequent phrases are strongly human-centric and appearance-centric, such as {\it man woman}, {\it her head}, {\it young man}, {\it her hands}, and several clothing- or body-related phrases. This indicates that the event captions in MEV are not only action descriptions, but are also consistently grounded in visible subjects, body parts, and local visual details. Figure~\ref{fig:sup_stats}(b) summarizes the most common action lemmas. High-frequency verbs such as {\it look}, {\it wear}, {\it hold}, {\it sit}, {\it smile}, and {\it stand} suggest that MEV covers a broad spectrum of event types, ranging from explicit physical actions to posture changes, gaze, and interaction-heavy moments that are common in long-take videos.

Figure~\ref{fig:sup_stats}(c) shows video-level camera-motion coverage: 51.1\% of videos contain at least one moving-camera event, and 59.7\% contain at least one static event. These categories overlap because a video can contain both types of events. At the event level, moving-camera and static events account for 42.9\% and 57.1\%, respectively. The two categories are relatively balanced. Figure~\ref{fig:sup_stats}(d) plots event duration for indices 1--17. Within this range, early events exhibit the largest spread and longest outliers, while later events are typically shorter and less variable. Intuitively, early events in long-take videos often serve as setup or scene establishment, whereas later events are more likely to be localized transitions or follow-up actions. The sparser points at later event indices also reflect the long-tailed distribution of event counts per video, with relatively few videos containing many events.

\noindent \textbf{Quality-score granularity.}
Across 8,016 videos and 25,262 events, we measure the share of total event-score variation that occurs among events within the same video. For each quality metric, we compute
\begin{equation}
R_{\mathrm{within}} = \frac{\sum_v\sum_e (x_{ve}-\bar{x}_v)^2}{\sum_v\sum_e (x_{ve}-\bar{x})^2},
\end{equation}
where $x_{ve}$ is the score of event $e$ in video $v$, $\bar{x}_v$ is that video's mean event score, and $\bar{x}$ is the mean across all events. Each event is weighted equally; Table~\ref{tab:quality_score_granularity} reports $100R_{\mathrm{within}}$.

\begin{table}[htbp]
\centering
\caption{\small \textbf{Within-video variation in event-level quality scores.} The percentage of total event-score variation occurring within videos is reported for each metric.}
\label{tab:quality_score_granularity}
\small
\setlength{\tabcolsep}{7pt}
\begin{tabular}{lr}
\toprule
\textbf{Metric} & \textbf{Within-video share (\%)} \\
\midrule
Subject consistency & 41.76 \\
Background consistency & 42.92 \\
Temporal flickering & 15.13 \\
Motion smoothness & 18.46 \\
Imaging quality & 11.39 \\
Aesthetic quality & 10.38 \\
\bottomrule
\end{tabular}
\end{table}

Subject and background consistency show substantial variation within videos, supporting event-level scoring for localized analysis. Imaging and aesthetic quality vary mainly between videos, supporting their use for global screening, while their event-level scores can still identify localized degradation.

\begin{figure*}[t]
  \centering
   \includegraphics[width=1.0\linewidth]{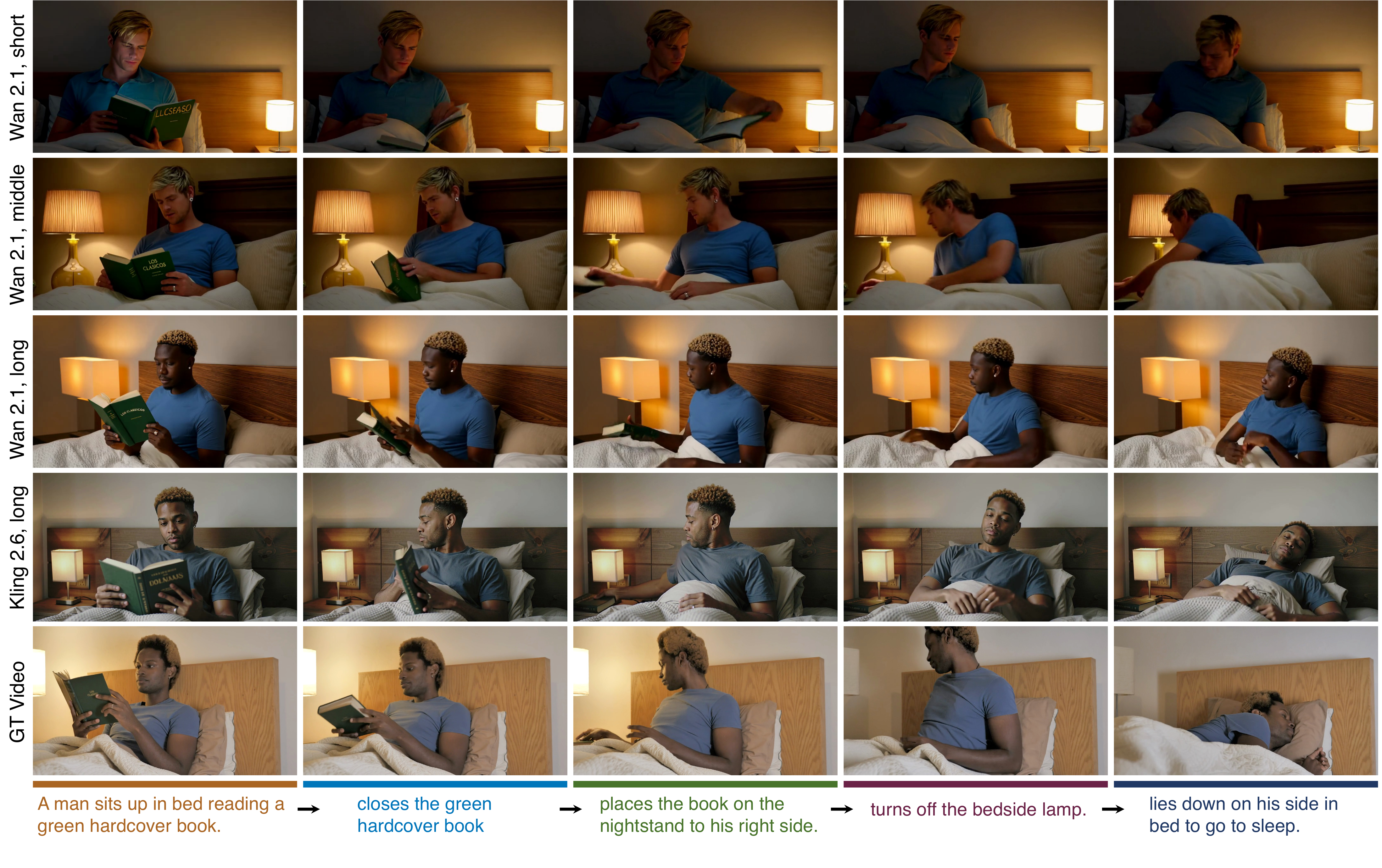}
   \caption{\small \textbf{Comparison of video generation performance across different caption granularities and models.} We evaluate the open-source Wan 2.1-14B~\citep{wan2025wan} and the closed-source Kling 2.6~\citep{kling2025video26} by concatenating global captions with event-level captions in a single-pass generation manner. Results show that higher caption granularity improves alignment with the Ground Truth video and enhances coverage of sequential events. Wan 2.1 consistently fails to capture some events (\eg, turning off the lamp and lying down), and even the closed-source Kling 2.6 struggles to execute ``turning off the lamp'', indicating potential limitations for multi-event video generation.}
   \label{fig:caption_granularity}
\end{figure*}

\noindent \textbf{Multi-granular caption quality analysis.}
Figure~\ref{fig:caption_granularity} illustrates the impact of global captions at different granularities (short/middle/long) on video generation. As the granularity becomes finer, the generative model can leverage more information. For example, the Middle Caption (row 2) correctly captures the person's orientation and preserves finer details (\eg, earrings and rings); the Long Caption (row 3) further specifies attributes such as the person's skin tone. Kling (row 4) achieves the best results, exhibiting a more faithful profile and more accurate event execution; however, it still omits certain events (\eg, ``turning off the lamp''). These observations suggest a potential limitation of current video generators: when handling multi-event video generation, some event-level information can be difficult to utilize effectively.
Video--text retrieval is an important application of video--language alignment~\citep{luo2022clip4clip,ma2022xclip,wang2023internvid,zhu2026mova}.
For retrieval, we fine-tune CLIP4Clip~\citep{luo2022clip4clip}, initialized from pretrained CLIP ViT-B/16, on MEV following its ActivityNet Captions~\citep{krishna2017dense} training configuration. We evaluate video--text retrieval using whole videos, with the same 80/20 split as the generation experiments (Appendix~\ref{sup:mevg}). For each setting, we concatenate the specified text fields from the same whole video into a single text input. Table~\ref{tab:multi_granularity_ablation} shows that enriching the short-caption baseline with global or event-level signals, including event captions, event camera descriptions, and global camera descriptions, consistently improves bidirectional retrieval performance.

\begin{table}[!h]
  \centering
  \caption{\small \textbf{Bidirectional video--text retrieval across multi-granular captions} (R@1).}
  \label{tab:multi_granularity_ablation}
  \setlength{\tabcolsep}{6pt}
  \begin{tabular}{lcc}
    \toprule
    Setting & T2V & V2T \\
    \midrule
    Base (Short) & 81.8 & 84.1 \\
    + Event camera desc & 84.8 & 85.2 \\
    + Event captions & 87.0 & 89.5 \\
    + Global camera & 83.8 & 87.4 \\
    \shortstack[l]{+ Global camera +\\ ~~~Event captions} & 85.9 & 86.9 \\
    \bottomrule
  \end{tabular}
  \vspace{-0.4em}
\end{table}

\noindent \textbf{Annotation model evaluation.}
We evaluate the original predictions of three VLM annotators against human annotations on event clips from 128 randomly sampled source videos (Table~\ref{tab:manual_attr_acc}). Camera motion is measured by exact-set accuracy: a prediction is correct only when its complete set of camera-motion labels matches the human annotation. Missing, additional, or directionally incorrect labels make the prediction incorrect. Semantic attributes are evaluated using binary accuracy. The main takeaway is that camera understanding remains the weakest component: although the models often recognize that the viewpoint changes, they still struggle to assign the correct fine-grained motion label. By contrast, most semantic attributes are substantially more reliable overall. Multi-person, animal, and cooking cues are comparatively easy, while lighting-related judgments remain noticeably less stable. This gap between camera precision and semantic robustness is consistent with the qualitative results below.

\begin{table*}[t]
\centering
\caption{\small \textbf{Accuracy (\%) of original VLM predictions against human annotations on event clips from 128 randomly sampled source videos.} Camera motion uses exact-set accuracy; semantic attributes use binary accuracy. Avg. is the unweighted mean of the eight displayed scores. Higher is better.}
\setlength{\tabcolsep}{4pt}
\renewcommand{\arraystretch}{1.03}
\scalebox{0.92}{
\begin{tabular}{@{}p{0.27\linewidth}*{5}{>{\centering\arraybackslash}p{0.116\linewidth}}@{}}
\toprule[1.0pt]
\textbf{Model} & \textbf{Camera} & \textbf{Appear.} & \textbf{Disappear.} & \makecell[c]{\textbf{Multi-} \\ \textbf{person}} & \makecell[c]{\textbf{Env.} \\ \textbf{Change}} \\
\midrule
\textbf{LLaVA-Video-7B} & 6.2 & 81.1 & 80.1 & 86.5 & 58.6 \\
\textbf{Gemini 2.5 Pro} & 22.0 & 91.5 & 89.8 & 95.5 & 76.6 \\
\textbf{Qwen3-VL-32B} & 10.0 & 91.3 & 89.6 & 96.7 & 72.8 \\
\specialrule{0.8pt}{0pt}{0pt}
\textbf{Model} & \textbf{Lighting} & \textbf{Animal} & \textbf{Cooking} & \multicolumn{2}{c}{\textbf{Avg.}} \\
\midrule
\textbf{LLaVA-Video-7B} & 55.6 & 89.4 & 90.5 & \multicolumn{2}{c}{68.5} \\
\textbf{Gemini 2.5 Pro} & 69.5 & 96.7 & 95.3 & \multicolumn{2}{c}{79.6} \\
\textbf{Qwen3-VL-32B} & 66.4 & 95.3 & 94.6 & \multicolumn{2}{c}{77.1} \\
\bottomrule[1.0pt]
\end{tabular}
}
\renewcommand{\arraystretch}{1.0}
\label{tab:manual_attr_acc}
\end{table*}

We additionally evaluate camera-specialized models from CameraBench~\citep{lin2025towards} using its released Qwen2.5-VL-7B and Qwen2.5-VL-72B checkpoints on 330 available event clips from the same 128 human-annotated source videos. We map the reference annotations and model predictions to CameraBench's 15 camera-motion outputs. Exact-set accuracy and F1 use a fixed Yes/No threshold of 0.5, without tuning on this subset. We compute mAP from continuous $P(\mathrm{Yes})$ scores. Table~\ref{tab:camerabench_mev} summarizes the results.

\begin{table}[htbp]
\centering
\caption{\small \textbf{Evaluation of CameraBench models on MEV (\%).} Results use 330 event clips from 128 source videos after mapping predictions and reference annotations to CameraBench's 15 camera-motion outputs. Higher is better.}
\label{tab:camerabench_mev}
\small
\setlength{\tabcolsep}{6pt}
\renewcommand{\arraystretch}{1.08}
\begin{tabular}{lrrrr}
\toprule
\textbf{Model} & \textbf{Exact-set} & \textbf{Micro-F1} & \textbf{Macro-F1} & \textbf{mAP} \\
\midrule
CameraBench-7B & 10.91 & 35.72 & 22.12 & 27.88 \\
CameraBench-72B & 8.18 & 34.70 & 24.53 & 24.13 \\
\bottomrule
\end{tabular}
\end{table}

Exact-set accuracy requires all motion labels in an event to be correct, whereas per-primitive AP evaluates each label independently. Among the 7B checkpoint's incorrect predictions, 46.3\% contain only omissions, with no additional incorrect labels. For 72B, the largest error category (46.2\%) combines omitted and additional incorrect labels. These patterns illustrate the difficulty of recovering complete label sets for composite camera motions.

MEV also annotates focus shift, which is not included in CameraBench's output taxonomy. For events with only a focus shift, the reference in the shared motion-label space is static; the focus change itself is not evaluated by these models. Overall, the specialized models still make substantial errors on MEV event clips, supporting the need for human verification in our annotation pipeline.

\noindent \textbf{Human verification and correction.}
Human annotators reviewed the camera labels and semantic attributes of all 25,262 events and corrected errors before release. Across the dataset, 83.2\% of Gemini-generated camera-label sets required at least one human correction. This rate counts events whose camera-label sets were changed, rather than individual label edits. It is computed over the full dataset, whereas the accuracy results in Table~\ref{tab:manual_attr_acc} use the 128-video evaluation subset.

\begin{figure*}[t]
    \centering
    \includegraphics[width=1.0\linewidth]{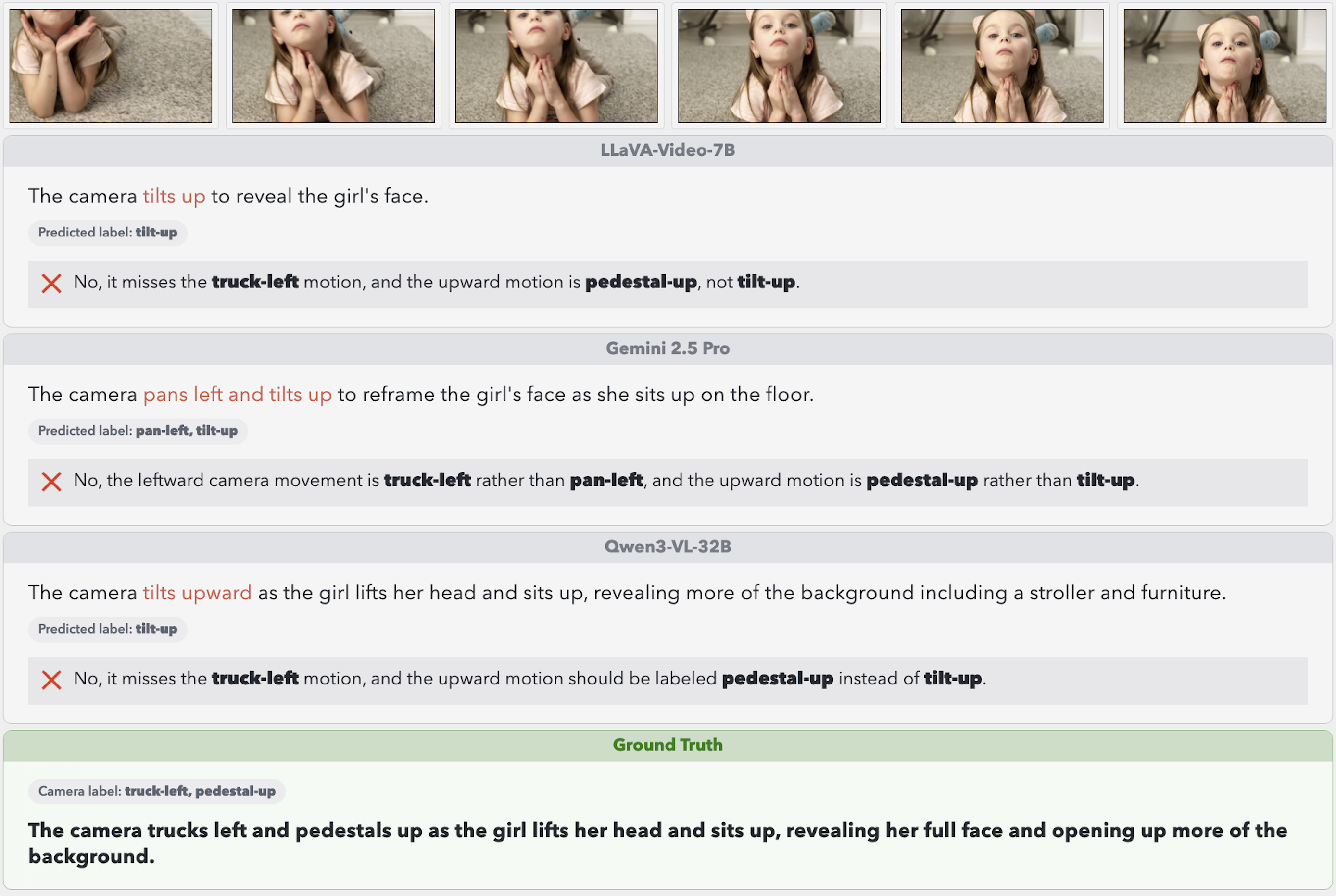}
    \caption{Qualitative comparison of camera motion understanding for different VLMs.}
    \label{fig:quali_camera}
\end{figure*}

\noindent \textbf{Qualitative analysis and failure taxonomy of camera annotations.}
Figure~\ref{fig:quali_camera} illustrates a representative failure mode of current VLMs on camera reasoning. In this example, all models capture that the framing changes as the girl lifts her head, but they fail to recover the correct motion decomposition. The ground-truth motion is a combination of {\tt truck-left} and {\tt pedestal-up}, whereas the models instead reduce the motion to {\tt tilt-up}, or further confuse translation with rotation by predicting {\tt pan-left}. This suggests that the challenge is not merely to notice motion, but to factorize it correctly into atomic primitives. Distinguishing physical camera translation from apparent framing rotation remains especially difficult.

\begin{figure*}[t]
    \centering
    \includegraphics[width=1.0\linewidth]{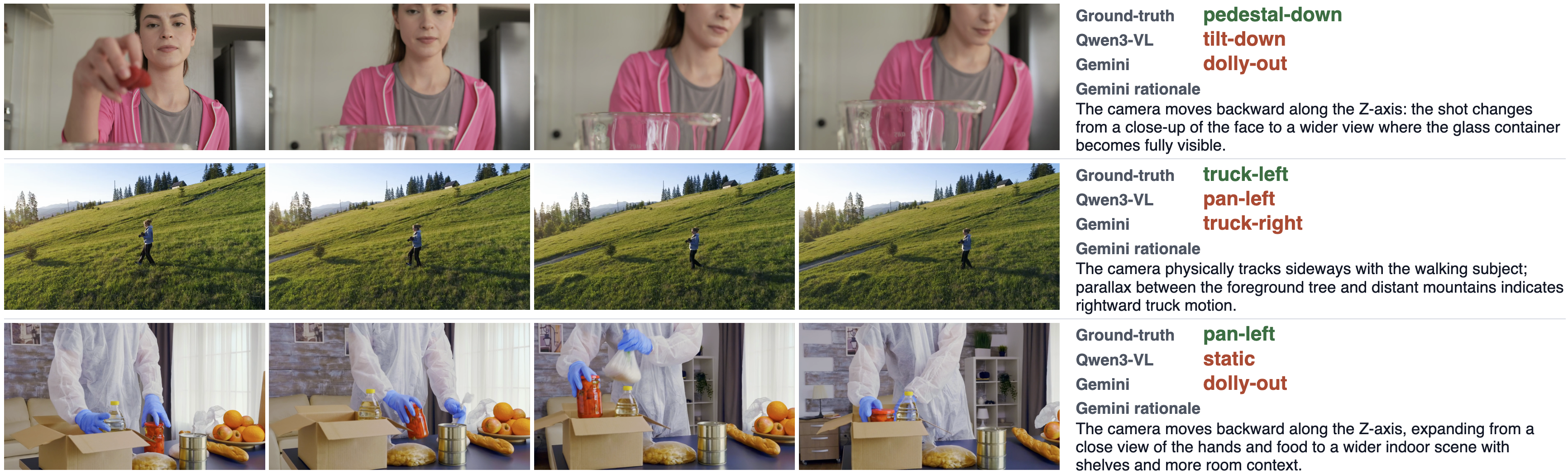}
    \caption{\small \textbf{Representative camera annotation failures.} Qwen3-VL often confuses camera rotation with physical translation, such as {\tt pan} versus {\tt truck} or {\tt tilt} versus {\tt pedestal}. Gemini 2.5 Pro gives richer 3D motion rationales, but can still misidentify the correct axis or direction, indicating that fine-grained camera annotation remains challenging for current VLMs.}
    \label{fig:camera_annotation_failure}
\end{figure*}

Figure~\ref{fig:camera_annotation_failure} further summarizes typical camera-label errors made by strong VLMs. Qwen3-VL is often ambiguous between rotation and translation, such as confusing {\tt pan} with {\tt truck}, or reducing vertical camera translation ({\tt pedestal}) to viewpoint rotation ({\tt tilt}). Gemini 2.5 Pro is less prone to some {\tt pan}/{\tt truck} confusions and often provides more detailed 3D motion rationales, but it can still assign the wrong motion axis or direction, such as predicting {\tt dolly-out} for cases whose visual change is better explained by {\tt pedestal-down} or {\tt pan-left}. Overall, these examples suggest that existing VLMs can recognize that the view changes, but their 3D scene understanding is still insufficient for reliably decomposing camera motion into translation, rotation, depth motion, and direction. This motivates our manual verification of camera annotations and the release of both categorical camera labels and natural-language camera descriptions.

\begin{figure*}[t]
    \centering
    \includegraphics[width=1.0\linewidth]{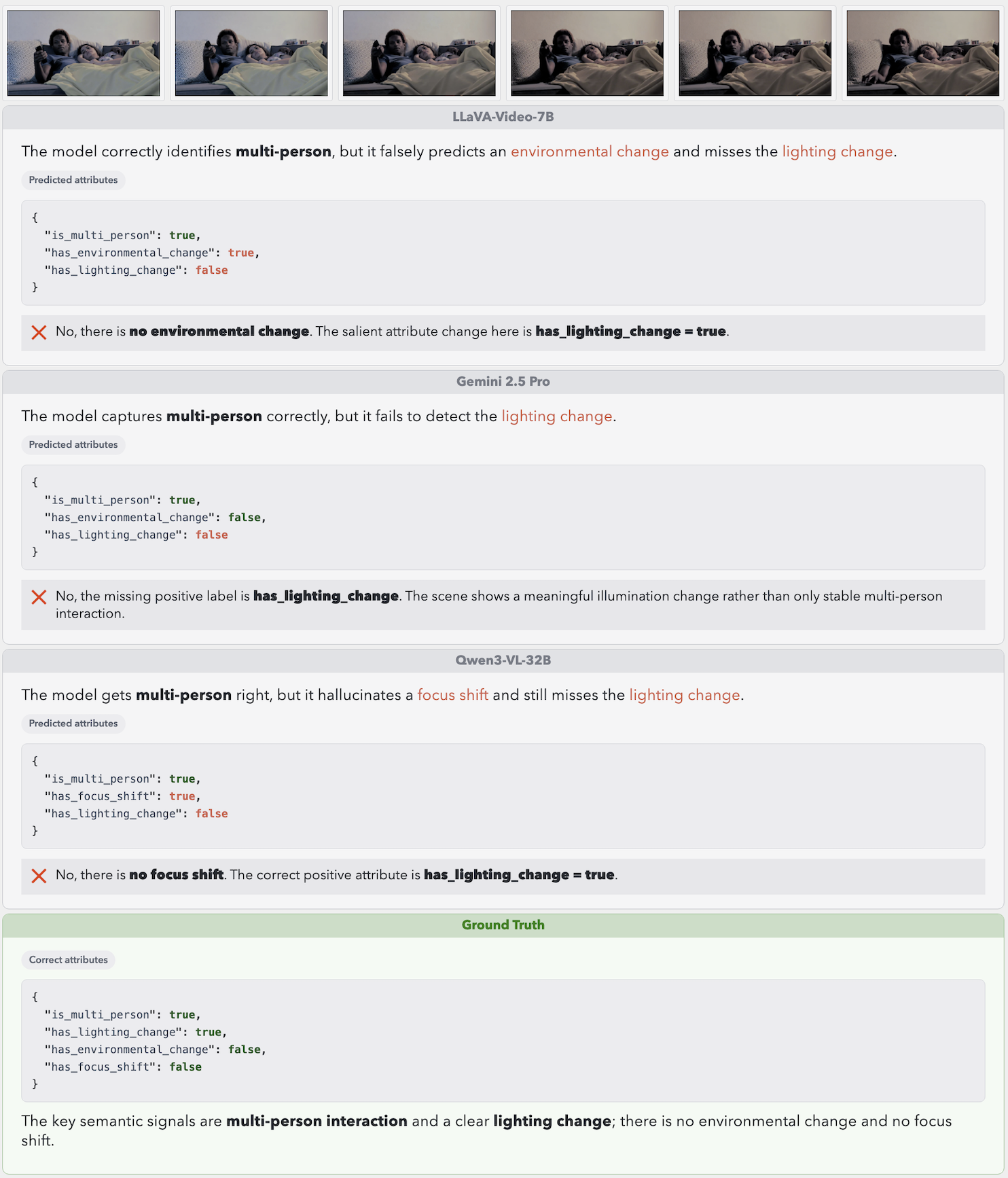}
    \caption{Qualitative comparison of semantic attribute prediction for different VLMs.}
    \label{fig:quali_semantic}
\end{figure*}

\noindent \textbf{Qualitative analysis of semantic attributes.}
Figure~\ref{fig:quali_semantic} shows that semantic attributes are easier than camera labels, but still not uniformly solved. The focus-shift predictions shown here are binary (\texttt{true}/\texttt{false}) judgments of whether a focus shift occurs. All three models correctly recognize the salient multi-person interaction, which is visually strong and semantically explicit. However, the lighting change is missed by all of them, indicating that illumination-related reasoning is still weak even when the scene content itself is understood. In addition, LLaVA hallucinates an environmental change, while Qwen3-VL-32B hallucinates a focus shift. These errors suggest that subtle visual changes are often assigned to the wrong underlying factor, especially when lighting, scene state, and photographic effects may be confounded.

\begin{figure*}[t]
    \centering
    \includegraphics[width=1.0\linewidth]{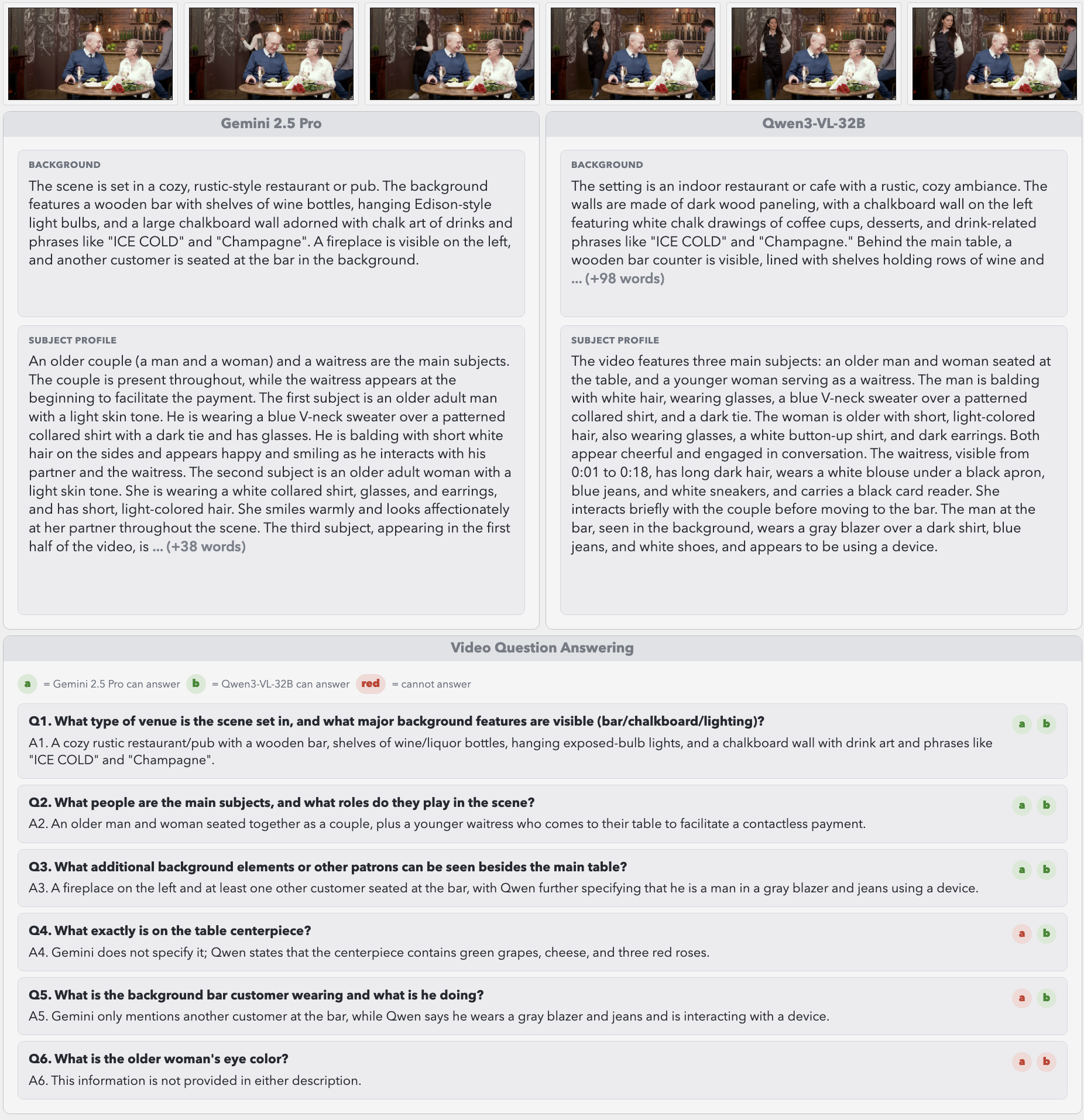}
    \caption{Global structured caption evaluation via video question-answering tasks. Color markers show answer coverage. Gemini produces a more compact description, while Qwen captures additional background details. Both omit some fine-grained attributes.}
    \label{fig:quali_global}
\end{figure*}

\noindent \textbf{Qualitative analysis of global structured captions.}
Figure~\ref{fig:quali_global} compares Gemini 2.5 Pro and Qwen3-VL-32B under the same prompting format for global structured captioning and downstream video question answering. A clear pattern is that the two models emphasize different aspects of the same scene. Gemini is more concise and better at producing a compact description that covers most of the useful global information with little redundancy. Qwen, in contrast, tends to be richer in background details and object-level specification. This difference becomes visible in the QA examples: both models answer coarse scene-level questions well, but Qwen is better at detail-sensitive questions such as the centerpiece composition or the clothing and activity of the background customer. At the same time, both models fail on very fine-grained details such as the older woman's eye color, which shows that even strong structured captions do not imply complete access to subtle visual attributes. Although Qwen captures more details relevant to the questions in this example, we select Gemini for MEV annotation based on its action-focused event descriptions and stronger overall performance in camera-motion and semantic-attribute annotation (Table~\ref{tab:manual_attr_acc}).

These analyses suggest a consistent hierarchy of difficulty on MEV. Coarse global understanding and object- or interaction-centric semantic attributes are already fairly accessible to modern VLMs. Fine-grained scene details can also be recovered, but model behavior becomes style-dependent: Gemini tends to be compact and efficient, while Qwen often provides denser background grounding. In contrast, exact camera-motion factorization and subtle lighting-related reasoning remain notably challenging. Our additional manual verification confirms that these weak points are not merely anecdotal captioning errors, but recurring limitations of current VLMs on structured long-take video analysis. In this sense, MEV helps reveal where existing models remain unreliable, while its human-verified annotations also make it easier to connect these observations to downstream tasks such as video captioning, event understanding, and long-take video generation.

\section{Implementation Details of MEV-Wan}
\label{sup:mevg}

\noindent \textbf{Overview.}
MEV-Wan is built on top of Wan2.1-T2V-1.3B and follows the MinT-style design for multi-event video generation. The key idea is to preserve Wan's pretrained single-video generation ability while adding explicit event-aware temporal conditioning. Concretely, we keep the original global text conditioning pathway for whole-video semantics and introduce an additional temporal conditioning pathway that aligns event captions with different time regions. The resulting model is trained with the same flow-matching objective as Wan, so the main change lies in the conditioning design rather than the denoising loss itself.

\noindent \textbf{Relationship among compared methods.}
All four methods belong to the Wan2.1-1.3B lineage. Wan and LongLive are evaluated using their released checkpoints without additional fine-tuning by us. Ultra-Wan and MEV-Wan use the same Wan backbone and MinT-style multi-event adaptation, with different training resources. Table~\ref{tab:method_comparison} summarizes these distinctions.

\begin{table}[htbp]
\centering
\caption{\small \textbf{Relationship among the four compared methods.} ``Fine-tuned by us'' refers to additional training performed in this work. A dash indicates that no additional fine-tuning data are used by us.}
\label{tab:method_comparison}
\small
\setlength{\tabcolsep}{4pt}
\renewcommand{\arraystretch}{1.15}
\begin{tabular}{@{}>{\raggedright\arraybackslash}p{0.16\linewidth}>{\raggedright\arraybackslash}p{0.13\linewidth}>{\raggedright\arraybackslash}p{0.24\linewidth}>{\raggedright\arraybackslash}p{0.39\linewidth}@{}}
\toprule
\textbf{Method} & \textbf{Fine-tuned by us?} & \textbf{Fine-tuning data used in this work} & \textbf{Multi-event mechanism} \\
\midrule
Wan2.1-1.3B & No & -- & Original Wan conditioning \\
\addlinespace
LongLive & No & -- & Official autoregressive checkpoint with prompt switching \\
\addlinespace
Ultra-Wan & Yes & UltraVideo-Long subset & MinT-style adaptation; inferred events and uniform boundaries \\
\addlinespace
MEV-Wan & Yes & MEV & MinT-style adaptation; verified events and grounded boundaries \\
\bottomrule
\end{tabular}
\end{table}

\noindent \textbf{Ultra-Wan data construction.}
We use temporal discourse connectives in UltraVideo-Long's Summarized Description field, such as ``then,'' ``after,'' and ``before,'' to infer the number of events $N$. Videos identified as multi-event are uniformly partitioned into $N$ segments and converted into MEV-style metadata. We select an UltraVideo-Long subset containing the same number of videos as MEV and fine-tune Wan with the same MinT-style multi-event adaptation used for MEV-Wan.

\noindent \textbf{Dual-path conditioning architecture.}
Each Wan diffusion transformer block retains the original global text cross-attention branch and augments it with a temporal cross-attention branch. The global branch receives the video-level middle caption and is responsible for preserving scene layout, appearance consistency, and overall style. The temporal branch receives the ordered event captions of the sampled event group and injects local event semantics into temporally aligned frame regions. Following MinT, the temporal branch is merged back through a bounded residual gate with {\tt Tanh} activation, and the gate is initialized near zero. This near-zero initialization is important in practice because it prevents the newly inserted branch from disrupting Wan's pretrained denoising trajectory too aggressively at the beginning of training.

For temporal binding, we adopt ReRoPE as in MinT. Instead of hard-assigning each event to a fully isolated frame segment, ReRoPE introduces a soft temporal bias based on event identities, normalized timestamps, event boundaries, and validity masks. This design is better suited to MEV because neighboring events in long-take videos often transition smoothly and may visually overlap near event boundaries.

\noindent \textbf{Dataset split.}
We partition the 8,016 original videos into training and test sets before constructing event groups. All event segments and contiguous groups derived from a given original video inherit its split assignment. Table~\ref{tab:mev_split} reports the numbers of original videos and annotated events, together with their total annotated event duration. The 80/20 ratio applies to original videos; event counts and durations need not follow exactly the same ratio.

\begin{table}[htbp]
\centering
\caption{\small \textbf{MEV training and test split statistics.} Duration is summed over annotated event intervals and is reported as HH:MM:SS.mmm. The split is performed at the original-video level before event-group construction.}
\label{tab:mev_split}
\setlength{\tabcolsep}{6pt}
\renewcommand{\arraystretch}{1.08}
\begin{tabular}{lrrr}
\toprule
\textbf{Split} & \textbf{Original videos} & \textbf{Events} & \textbf{Annotated event duration} \\
\midrule
Train & 6,413 & 20,241 & 29:07:24.974 \\
Test & 1,603 & 5,021 & 07:48:42.790 \\
\midrule
Total & 8,016 & 25,262 & 36:56:07.764 \\
\bottomrule
\end{tabular}
\end{table}

\noindent \textbf{Cross-split overlap audit.}
We extract DINOv2 features~\citep{oquab2023dinov2} from frames sampled at 1 fps to identify potential visual overlap between the training and test sets. A video pair is flagged when at least 80\% of the sampled frames in the shorter video have temporally ordered matches in the longer video with cosine similarity $\geq 0.95$. All flagged pairs are manually reviewed. No confirmed duplicate or near-duplicate pairs are found under this protocol. The two splits may share source platforms, such as MixKit and YouTube, while all events derived from the same original video remain within one split.

\noindent \textbf{Training sample construction.}
Suppose a video contains ordered events $\{e_i\}_{i=1}^{N}$ with annotated durations. We construct MEV-Wan training samples using \emph{contiguous event grouping} by default. More specifically, we slide a window over adjacent events and sample groups containing 2 to 5 events. This matches the downstream generation setting, where the model is expected to synthesize a temporally continuous multi-event clip rather than an unordered collection of independent actions.

After selecting an event group, we allocate training frames to events proportionally to their durations. If the selected durations are $\{d_i\}$ and the training clip contains $T$ frames, the frame budget for event $i$ is approximately $T \cdot d_i / \sum_j d_j$, followed by a rounding-and-correction step to ensure that the allocated frame counts sum to $T$. This duration-aware construction yields frame-wise event indices, normalized timestamps, event validity masks, and aligned temporal spans for the temporal branch. Compared with naively concatenating event captions, this step teaches the model not only \emph{what} events should happen, but also \emph{when} they should appear and \emph{for how long} they should persist.

\noindent \textbf{Optimization and training procedure.}
MEV-Wan is optimized with the standard Wan flow-matching supervised fine-tuning objective. In each iteration, we sample a video clip, select a contiguous event group, construct the duration-aware temporal tensors, encode the clip into Wan latents, sample a flow-matching timestep, and train the denoiser to predict the target velocity under both global and temporal conditions. In this sense, MEV-Wan reuses the stable Wan training pipeline while extending the conditioning space to become explicitly multi-event aware.

Our implementation supports both full fine-tuning and LoRA fine-tuning under a unified code path. Consistent with our default setting, MEV-Wan uses LoRA-based adaptation for efficient multi-event training, while full fine-tuning of the Wan {\tt dit} module is retained only as an optional implementation path. In the default LoRA setup, we use rank 32 and adapt the standard Wan attention and feed-forward projections, namely {\tt q,k,v,o,ffn.0,ffn.2}.

\noindent \textbf{Implementation choices.}
Several engineering choices are important for stable multi-event adaptation. First, we cap the number of events per training sample at 5 and use explicit padding masks to support variable-length event groups without changing the architecture. Second, we keep gradient checkpointing enabled and use memory-oriented runtime settings to reduce fragmentation and peak memory overhead for long video clips. Finally, the combination of near-zero temporal gate initialization and default LoRA-based adaptation helps reduce catastrophic forgetting of Wan's original generation ability.

\begin{table}[t]
\centering
\caption{\small \textbf{Default training hyperparameters for MEV-Wan on MEV.} We list the parameters that directly determine the model configuration or the multi-event adaptation behavior.}
\label{tab:wan_mev_hparams}
\resizebox{\linewidth}{!}{
\begin{tabular}{l l l}
\toprule[1.2pt]
\textbf{Category} & \textbf{Hyperparameter} & \textbf{Value} \\
\midrule
Video setup & Backbone & Wan2.1-T2V-1.3B \\
Video setup & Resolution & $480 \times 832$ \\
Video setup & Number of frames & 81 \\
\midrule
Optimization & Optimizer & AdamW \\
Optimization & Learning rate & $1\times10^{-5}$ \\
Optimization & Weight decay & 0.01 \\
Optimization & LR schedule & Constant \\
Optimization & Per-GPU batch size & 1 video \\
Optimization & Effective total batch size & 128 videos \\
Optimization & Number of epochs & 8 \\
\midrule
Multi-event specific & Event count per sample & 2--5 \\
Multi-event specific & Grouping policy & Contiguous sliding window \\
Multi-event specific & Frame allocation & Duration-aware proportional allocation \\
Multi-event specific & ReRoPE rescale length $L$ & 8 \\
Multi-event specific & Temporal gate init & Near zero \\
Multi-event specific & Scene-cut conditioning & Disabled \\
\midrule
LoRA & LoRA rank & 32 \\
LoRA & LoRA target modules & {\tt q,k,v,o,ffn.0,ffn.2} \\
\bottomrule[1.2pt]
\end{tabular}
}
\end{table}

\noindent \textbf{Compute resources.}
The reported experiments are conducted on AMD MI210 GPUs. Across the MEV-Wan training runs, annotation-model training, generation, and evaluation used for the reported results, the total compute budget is approximately 2{,}800 GPU-hours. Unless otherwise stated, training uses a per-GPU batch size of 1 and an effective total batch size of 128. For generation evaluation, each 128-prompt test set is randomly sampled five times from the MEV test split, and we report the average over the five draws.

\section{MEV-Eval Protocol for Multi-Event Video Generation}
\label{sup:mev_eval}

\noindent \textbf{Protocol overview.}
MEV-Eval is an event-centric evaluation protocol designed to complement general video-quality metrics such as VBench. Given a generated video and an input event sequence $E=\{e_i\}_{i=1}^{N}$ with duration assignments $\{\tau_i\}_{i=1}^{N}$, MEV-Eval asks whether the video executes the intended events, preserves their order, grounds them to the intended temporal regions, and transitions coherently between adjacent events. In our default implementation, we use Gemini 2.5 Pro as the VLM judge. The target temporal assignments are derived from MEV's human-verified event annotations. Gemini evaluates the generated video against these fixed targets. The judge is provided with the generated video, the global prompt, the ordered event captions, and the normalized duration assignment of each event. Method identities are hidden from the judge, and all methods are evaluated with the same event list, video format, scoring rubric, and output schema.

\noindent \textbf{Event Execution Accuracy (EEA).}
For each event $e_i$, the judge outputs a binary label $x_i\in\{0,1\}$ indicating whether the event is visually and semantically executed. An event is counted as executed only when the relevant action and key subject/object are visible; missing, wrong, or only vaguely implied events are counted as failures. We compute
\begin{equation}
\mathrm{EEA}=\frac{1}{N}\sum_{i=1}^{N}x_i .
\end{equation}

\noindent \textbf{Event Order Accuracy (EOA).}
The judge identifies the observed order of executed events. We then compute pairwise order accuracy over all event pairs:
\begin{equation}
\mathrm{EOA}=\frac{1}{\binom{N}{2}}\sum_{i<j}\mathbf{1}\left[e_i \prec e_j\right],
\end{equation}
where $e_i \prec e_j$ means that event $e_i$ appears before $e_j$ in the generated video. If either event in a pair is missing or cannot be localized reliably, that pair is counted as incorrect.

\noindent \textbf{Temporal Grounding Accuracy (TGA).}
The generated video is evaluated against the same duration-normalized temporal assignment used by the input event sequence. For each event region $\tau_i$, the judge outputs $g_i\in\{0,1\}$ indicating whether the corresponding temporal region mainly depicts event $e_i$. We compute
\begin{equation}
\mathrm{TGA}=\frac{1}{N}\sum_{i=1}^{N}g_i .
\end{equation}
This metric differs from EEA: a video can execute all events but still receive a low TGA score if the events appear in the wrong temporal regions or are mixed across multiple regions.

\begingroup
\displaywidowpenalty=10000
\noindent \textbf{Boundary Transition Coherence (BTC).}
For each adjacent event boundary, the judge rates the transition on a 1--5 scale based on semantic continuity, subject identity stability, object-state consistency, scene/camera continuity, and the absence of severe drift, stagnation, or object splitting. A score of 1 indicates a broken or unrelated transition, 3 indicates a partially coherent transition with visible inconsistency, and 5 indicates a smooth and logically coherent transition. We normalize the score to $[0,1]$:
\begin{equation}
\mathrm{BTC}=\frac{1}{5(N-1)}\sum_{i=1}^{N-1}b_i ,
\end{equation}
where $b_i$ is the 1--5 rating for the boundary between events $e_i$ and $e_{i+1}$.
\par
\endgroup

\noindent \textbf{Human validation.}
We use a fixed random sample of 128 test prompts. Each of the four methods generates one video per prompt, yielding 512 videos. Three annotators independently score these videos using the same EEA, EOA, TGA, and BTC criteria, with method identities hidden. We use these judgments to assess agreement with Gemini and to compare generation methods under human evaluation. For EEA and TGA, the decision unit is an event-level binary label. For EOA, the decision unit is a pairwise event-order label. For BTC, the decision unit is an adjacent-boundary 1--5 coherence score. For EEA, EOA, and TGA, we report agreement between the VLM label and the majority-vote human consensus, together with mean pairwise human inter-rater agreement. For BTC, we report Spearman rank correlation between the VLM score and the mean human score, together with within-one rating agreement among humans.

\begin{table}[t]
\centering
\caption{\small \textbf{Human validation of MEV-Eval judging.} These statistics validate the VLM-assisted judge and are not additional generation metrics. For EEA, EOA, and TGA, agreement is computed on binary decision units; for BTC, validation compares 1--5 boundary-transition ratings.}
\label{tab:mev_eval_human_validation}
\vspace{0.5em}
\setlength{\tabcolsep}{4pt}
\renewcommand{\arraystretch}{1.12}
\begin{adjustbox}{max width=\linewidth}
\begin{tabular}{lccc}
\toprule
Metric & Decision unit & VLM--human validation & Human inter-rater validation \\
\midrule
EEA & Event execution label & Agreement w/ consensus: 0.82 & Pairwise agreement: 0.86 \\
EOA & Event-pair order label & Agreement w/ consensus: 0.76 & Pairwise agreement: 0.81 \\
TGA & Event-region match label & Agreement w/ consensus: 0.73 & Pairwise agreement: 0.78 \\
BTC & Boundary coherence score & Spearman $\rho$: 0.64 & Within-1 agreement: 0.83 \\
\bottomrule
\end{tabular}
\end{adjustbox}
\vspace{-0.6em}
\end{table}

Using the same videos and human judgments, we also compare Gemini and human scores for each method in Table~\ref{tab:mev_eval_human_scores}. Human EEA, EOA, and TGA use the majority decision of the three annotators for each decision unit. For BTC, we average their 1--5 ratings for each boundary and apply the same normalization as above. We compute scores per prompt using the metric definitions above and then average over the 128 prompts. These results use one fixed subset, while Table~\ref{tab:mev_eval} reports averages over five random draws.

\begin{table}[htbp]
\centering
\caption{\small \textbf{Human and Gemini evaluation on the same generated videos.} Four methods are evaluated on 128 fixed test prompts, yielding 512 videos. The human judgments are the same as those used in Table~\ref{tab:mev_eval_human_validation}. Best scores for each judge are in bold.}
\label{tab:mev_eval_human_scores}
\small
\setlength{\tabcolsep}{7pt}
\renewcommand{\arraystretch}{1.08}
\begin{tabular}{llcccc}
\toprule
\textbf{Method} & \textbf{Judge} & \textbf{EEA}$\uparrow$ & \textbf{EOA}$\uparrow$ & \textbf{TGA}$\uparrow$ & \textbf{BTC}$\uparrow$ \\
\midrule
\multirow{2}{*}{Wan2.1-1.3B} & Gemini & 0.38 & 0.23 & 0.25 & 0.45 \\
 & Human & 0.42 & 0.19 & 0.20 & 0.47 \\
\addlinespace
\multirow{2}{*}{LongLive} & Gemini & 0.49 & 0.35 & 0.41 & 0.56 \\
 & Human & 0.53 & 0.29 & 0.34 & 0.60 \\
\addlinespace
\multirow{2}{*}{Ultra-Wan} & Gemini & 0.44 & 0.30 & 0.40 & 0.53 \\
 & Human & 0.46 & 0.31 & 0.36 & 0.56 \\
\addlinespace
\multirow{2}{*}{MEV-Wan} & Gemini & \textbf{0.65} & \textbf{0.45} & \textbf{0.58} & \textbf{0.60} \\
 & Human & \textbf{0.70} & \textbf{0.44} & \textbf{0.52} & \textbf{0.68} \\
\bottomrule
\end{tabular}
\end{table}

Both human and Gemini evaluations rank MEV-Wan first on all four metrics. The relative ordering of LongLive and Ultra-Wan changes for EOA and TGA, showing that judge differences affect some comparisons while preserving MEV-Wan's advantage on this subset.

\section{Additional Qualitative Results for Multi-Event Generation}
\label{sup:qual_gen}

\begin{figure*}[t]
    \centering
    \includegraphics[width=1.0\linewidth]{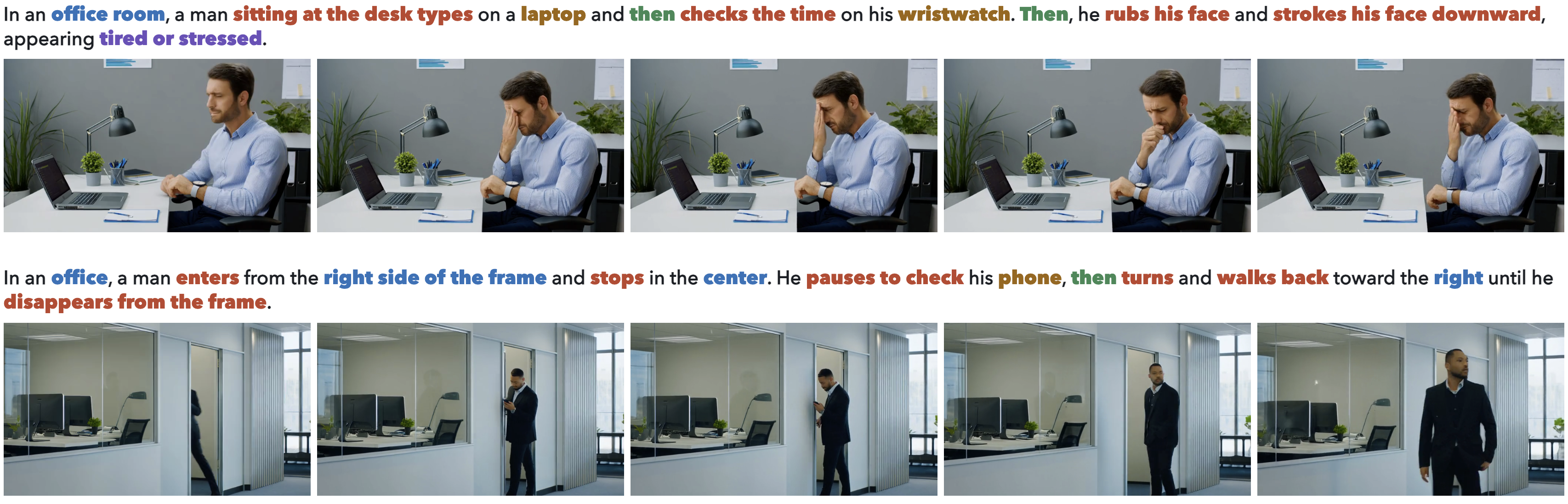}
    \caption{\small \textbf{Qualitative examples under flat text-only conditioning.} These Wan2.1-1.3B examples suggest that flat text-only conditioning can struggle to organize rich action cues into a coherent multi-event progression without structured event-level supervision.}
    \label{fig:bad_wo_boundary}
\end{figure*}

\noindent \textbf{Challenges under flat text-only conditioning.}
Figure~\ref{fig:bad_wo_boundary} provides qualitative examples from Wan2.1-1.3B under flat text-only conditioning. The prompts contain clear action information and temporal cues such as ``then'', but the generated videos do not consistently organize all instructed actions into a stable temporal progression. These cases suggest that multi-event video generation can be difficult when rich action descriptions are not paired with structured event-level conditioning. MEV addresses this setting by providing event captions and temporal boundaries that make the intended event sequence explicit during training.

\begin{figure*}[t]
    \centering
    \includegraphics[width=1.0\linewidth]{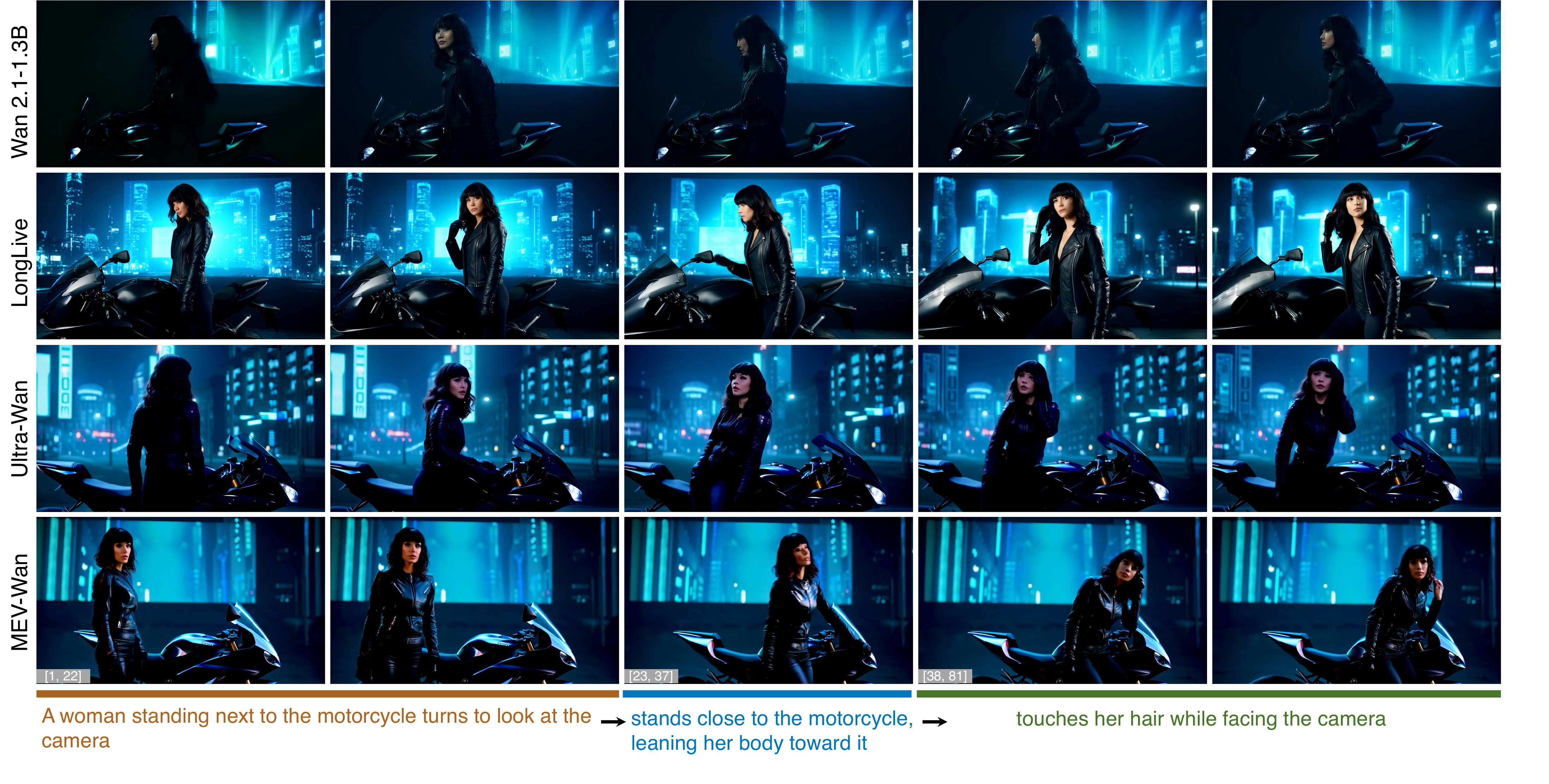}
    \caption{Additional qualitative comparison on multi-event video generation. }
    \label{fig:sup_gen}
\end{figure*}

\noindent \textbf{Generation comparison.}
Figure~\ref{fig:sup_gen} shows an additional qualitative comparison on a three-event generation example. Compared with the vanilla Wan2.1 baseline, both MEV-Wan and Ultra-Wan exhibit a clearer event structure and stronger instruction-following ability. The generated sequence more faithfully reflects the intended progression from turning to look at the camera, to leaning the body toward the motorcycle, and then to touching the hair while facing the camera. In contrast, the original Wan baseline preserves the overall scene style but is weaker at expressing the staged temporal structure specified by the prompt.
Among the compared methods, MEV-Wan provides the best alignment between event order and visual evolution. The pose change, body orientation, and interaction with the motorcycle remain more consistent with the three event descriptions across the full sequence. Ultra-Wan also improves over vanilla Wan, suggesting that explicit multi-event conditioning is helpful even in a relatively short three-event setting, although its cross-event continuity and instruction following are still weaker than MEV-Wan.
LongLive still suffers from severe drifting even when the prompt contains only three events. In particular, a bright patch appears on the right side of the last two frames, forming an obvious artifact that is unrelated to the target scene. This failure is accompanied by less stable subject control and weaker temporal consistency. Overall, this example suggests that event-aware conditioning improves both event structure and instruction following, while autoregressive long-video generation remains vulnerable to drift accumulation even in comparatively simple cases.

\section{Broader Use, Limitations, and Future Directions}
\label{sup:future_use}

\begin{figure*}[t]
    \centering
    \includegraphics[width=1.0\linewidth]{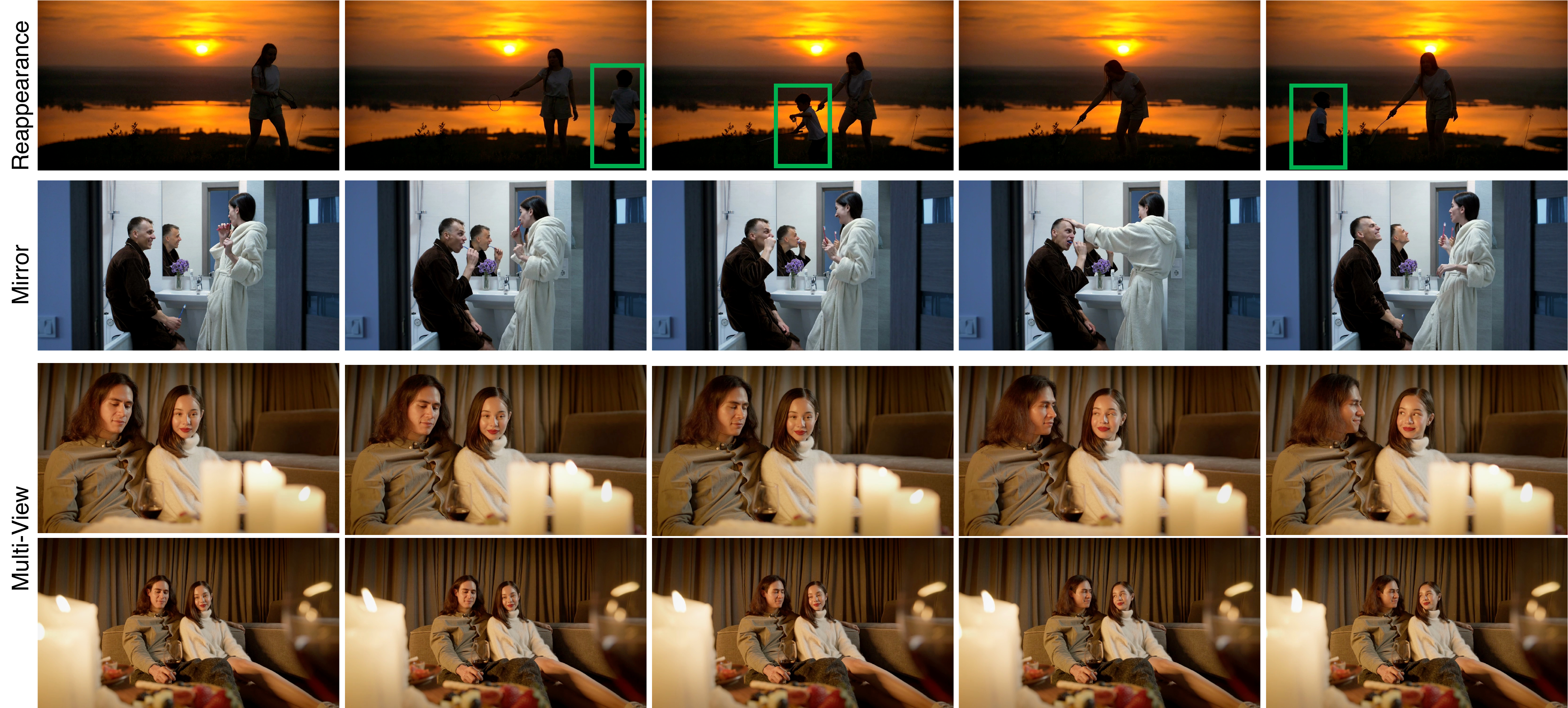}
    \caption{\small \textbf{Broader future use of MEV.} MEV contains challenging long-take cases involving reappearance, mirror reasoning, and multi-view consistency, which can support future attribute-guided studies and event-level analysis beyond the current generation experiments.}
    \label{fig:broader_use}
\end{figure*}

Beyond the attributes used in the current release, MEV also contains cases that are useful for studying more specialized long-take generation challenges. Figure~\ref{fig:broader_use} shows three representative directions. First, \textbf{reappearance} requires a model to preserve identity, spatial context, and object state when a subject temporarily disappears because of occlusion, framing, or scene layout and later returns to the visible region. This is challenging for current video generation models because the subject is no longer continuously observed, yet the later event must remain consistent with earlier context rather than introducing a new identity or duplicating the subject. Second, \textbf{mirror} scenes require reasoning about real and reflected spaces simultaneously. A model must keep body pose, object interaction, directionality, and reflection geometry consistent, while avoiding the common error of treating the mirror image as an independent actor or producing physically inconsistent reflected motion. Third, \textbf{multi-view} cases require stable event understanding across viewpoint changes, foreground occlusion, and perspective-dependent appearance. The model must preserve the same subjects and scene state while the camera or view configuration changes, which is difficult when generation is driven only by coarse text descriptions.

MEV's structured annotations support broader applications in video generation and understanding. Its event-level camera descriptions and semantic attributes support research on controllable video generation~\citep{he2025cameractrl,wang2024motionctrl,shen2026controllable}. The paired event captions and temporal boundaries support language-guided temporal grounding~\citep{lin2023univtg,wang2025timer1}, while the ordered event sequences and structured annotations support studies of event-to-event relations and causal discovery over visual state changes. Compared with video reasoning datasets such as MECD~\citep{chen2024mecd}, MEV provides a larger-scale video collection with richer generation-oriented conditioning signals and more explicit temporal changes, making it complementary for investigating how event transitions, visual state changes, and controllable conditioning interact in long videos.

\noindent \textbf{Broader impacts.}
MEV can support positive research progress in controllable long-video generation, event-level video understanding, and transparent dataset auditing by providing temporally grounded supervision rather than only coarse video--text pairs. At the same time, stronger long-video generation models may be misused to create misleading synthetic content or imitate recognizable public scenes, and public-video datasets can inherit source and content biases. We mitigate these risks by filtering sensitive and private-context videos, avoiding sensitive personal-attribute annotation, documenting data sources and licensing constraints, and supporting removal or modification of affected materials when rights, privacy, or safety concerns arise.

\noindent \textbf{Limitations and future directions.}
Extending the evaluation of MEV fine-tuning to other model scales and video domains remains future work.
Although we introduce event-centric criteria for multi-event video generation, such as event execution, event order, temporal grounding, and transition coherence, scalable evaluation still relies heavily on VLM-based judging. This remains imperfect because current VLMs can struggle with fine-grained temporal localization, subtle state changes, camera motion, and reflection or multi-view reasoning, as also indicated by our annotation-model analysis. Future work should therefore improve human validation, judge calibration, and uncertainty estimates for event-level video generation evaluation. In addition, our automatic annotation pipeline still depends on manual verification and correction for fields such as event captions, semantic attributes, and camera descriptions. More accurate annotation models, uncertainty-aware filtering, model-ensemble checks, and task-specific annotators for camera motion or hard attributes could reduce this correction burden while preserving the reliability required for temporally grounded multi-event supervision.

\clearpage

\end{document}